\documentclass[letterpaper, 10pt, conference]{ieeeconf}
\IEEEoverridecommandlockouts
\usepackage{graphicx}
\usepackage{caption}
\usepackage{amsmath, amsfonts}
\usepackage{cleveref}
\usepackage{booktabs}
\usepackage{tikz}
\usepackage{float}
\usepackage{siunitx}
\usepackage{subcaption}
\usepackage{pgfplots}
\pgfplotsset{compat=1.18} 
\usepackage[table]{xcolor}
\usepackage{mathtools} 
\usetikzlibrary{arrows.meta, positioning, shapes.geometric, fit}

\usetikzlibrary{shapes.geometric, arrows}
\tikzstyle{startstop} = [rectangle, rounded corners, minimum width=3.5cm, minimum height=1cm, text centered, draw=black]
\tikzstyle{process} = [rectangle, minimum width=3.5cm, minimum height=1cm, text centered, draw=black]
\tikzstyle{arrow} = [thick,->,>=stealth]

\begin{document}

\title {Reproducing and Extending RaDelft 4D Radar with Camera-Assisted Labels}


\author{
    K. Hu\textsuperscript{1},
    M. Alsakabi\textsuperscript{1}, 
    J. M. Dolan\textsuperscript{2},
    O. K. Tonguz\textsuperscript{1}, 
    \\
    \textsuperscript{1} Department of Electrical and Computer Engineering, College of Engineering \\
    \textsuperscript{2} The Robotics Institute, School of Computer Science \\
    Carnegie Mellon University, Pittsburgh, PA, United States \\
    \{kejiah, malsakab, tonguz, jdolan\}@andrew.cmu.edu, 
}

\maketitle

\begin{abstract}
Recent advances in 4D radar highlight its potential for robust environment perception under adverse conditions, yet progress in radar semantic segmentation remains constrained by the scarcity of open source datasets and labels. The RaDelft data set \cite{radelft_dataset}, although seminal, provides only LiDAR annotations and no public code to generate radar labels, limiting reproducibility and downstream research. In this work, we reproduce the numerical results of the RaDelft group \cite{sun2025automatic} and demonstrate that a camera-guided radar labeling pipeline can generate accurate labels for radar point clouds without relying on human annotations. By projecting radar point clouds into camera-based semantic segmentation and applying spatial clustering, we create labels that significantly enhance the accuracy of radar labels. These results establish a reproducible framework that allows the research community to train and evaluate the labeled 4D radar data. In addition, we study and quantify how different fog levels affect the radar labeling performance.
\end{abstract}

\begin{keywords}
Semantic Scene Understanding, Sensor Fusion, Deep Learning for Visual Perception, Computer Vision for Transportation, Data Sets for Robotic Vision
\end{keywords}

\IEEEpeerreviewmaketitle

\section{Introduction}
4D radar is increasingly used for autonomous perception due to its long-range sensing and robustness to adverse weather conditions. RaDelft data \cite{sun2025automatic} report excellent quantitative results, but do not release code or radar labels, making it difficult to validate methods, compare architectures fairly, or extend pipelines to new environments and settings. This reproducibility gap creates barriers for researchers who want to use the RaDelft dataset.

This paper addresses reproducibility and accessibility to labels. First, we reproduce the RaDelft radar semantic segmentation baseline without using the authors' code or proprietary labels, relying solely on publicly available radar tensors and calibration metadata. We implement a training and evaluation pipeline and release the complete package: code and trained checkpoints. Second, we propose a camera-guided radar labeling pipeline that transfers semantic priors from camera segmentation to radar, followed by class-aware clustering to refine the transferred labels. This cross-modal supervision significantly improves the detection probability compared to a radar-only setup. Third, we study adversarial weather robustness by simulating fog at varying intensities in the image domain and measuring how fog degrades image semantic segmentation and final radar labeling performance. This analysis clarifies how camera-based labeling benefits radar under degraded visual conditions.

Our main contributions are: (1) An open and reproducible baseline on RaDelft. The complete training and inference code, trained weights, and evaluation scripts will be released upon acceptance. (2) A practical pipeline that uses cameras to guide radar labeling. This produces labeled radar point clouds and achieves better detection probability and improved chamfer distance than the original RaDelft results \cite{sun2025automatic}. (3) We vary the intensity of the fog in the camera domain to measure its impact on camera-guided labeling, offering insights into cross-modal supervision under adverse weather.

\section{Related Work}
Deep learning has significantly advanced 3D semantic segmentation by enabling networks to learn spatial and contextual information from multi-modal data \cite{qi2017pointnet, qi2017pointnet++, guo2020deep}. The recent survey by Betsas \textit{et al.} provides a comprehensive overview of deep learning methods for 3D semantic segmentation \cite{betsas2025deep}. For example, Schumann \textit{et al.} use a variant of PointNet++ to perform semantic segmentation in radar point clouds \cite{schumann2018semantic}. However, 3D semantic segmentation suffers from the limitation of each sensor's capability. This motivates the use of camera segmentation as a supervision method for generating labeled point clouds. Yan \textit{et al.} introduced a 2D priors-assisted semantic segmentation approach that transfers rich semantic cues from camera image to LiDAR point clouds, which effectively improved 3D scene understanding \cite{yan20222dpass}. Sun \textit{et al.} used a similar method to label LiDAR point clouds in the RaDelft data set \cite{sun2025automatic}. This paradigm demonstrates the benefit of leveraging 2D priors to enhance 3D perception, which aligns with our goal of fusing camera semantics to help radar-based scene understanding. However, these works do not investigate how degraded environments, such as fog or low visibility, affect the accuracy of camera-based supervision. This gap motivates our study, which analyzes the robustness of camera-assisted labeling under different fog intensities and explores the complementary role of radar in adverse weather.
\section{Methodology}
This section outlines the main components and methodology for reproducing the RaDelft baseline, camera radar fusion, and camera-assisted labeling.
\subsection{Reproducing the RaDelft Baseline}
We follow the RaDelft convention and reproduce the results. 
\subsubsection{RAED $\rightarrow$ RAE encoding}
Each radar frame is first represented as a Range-Azimuth-Elevation-Doppler (RAED) tensor of size $2\times 128\times 240\times 500$ (channels $\times$ Doppler $\times$ azimuth $\times$ range). The two channels store (i) power measured at a specific Doppler, azimuth, and range bin, and (ii) a single elevation index $\in\{1,\ldots,34\}$ per RAD voxel corresponding to the strongest return along the elevation dimension for that voxel. For supervision and inference, we convert the RAED tensor to a Range--Azimuth--Elevation (RAE) volume of size $500\times 240\times 34$ using the provided elevation index and average-pool power aggregation within bins, consistent with the baseline formulation. Then, we apply $\log(1+p)$ and per-frame standardization over $(R,A)$ to stabilize the dynamic range:
\[
\tilde{p}=\log(1+p),\qquad
\hat{p}=\frac{\tilde{p}-\mu_{RA}}{\sigma_{RA}+\varepsilon}\,
\]
where $p$ is the raw power at each voxel, $\mu_{RA}$ and $\sigma_{RA}$ denote the mean and standard deviation of $\tilde{p}$ computed across the range–azimuth plane $(R,A)$ for each frame, ensuring that the normalized features are zero-centered and scale-invariant with each radar frame. The small constant $\varepsilon$ is added to avoid numerical instability during division.
This yields the RAE tensor for the 2D backbone.

\subsubsection{Backbone (two 2D FPN branches + 3D U-Net refinement)}
Following the RaDelft architecture design, we create two ResNet-18 FPN \cite{lin2017feature} branches operating on  $(N_E,N_R,N_A)$ slices: an \emph{occupancy} branch producing $O\in\mathbb{R}^{B\times N_E\times N_R\times N_A}$ (batch, elevation, range, azimuth) and a \emph{class} branch producing $C\in\mathbb{R}^{B\times K\times N_R\times N_A}$ (batch, classes, range, azimuth), $K$ = 5 in this work, i.e, 5 classes of
interest. Broadcasting these into a 3D semantic seed, 
\[
S_{\text{init}}=\sigma(O)\odot \mathrm{softmax}(C)\ \in\ \mathbb{R}^{B\times K\times N_R\times N_A\times N_E},
\]
where $\sigma$ represents the sigmoid function, $\odot$ denotes element-wise multiplication between two tensors of the same shape, and the softmax function converts a vector of class logits into a normalized probability distribution whose entries sum to 1.
We followed with a lightweight 3D U-Net \cite{cciccek20163d} which has two encoder levels, bottleneck, and symmetric decoder. The network outputs logits
$Z\in\mathbb{R}^{B\times K\times N_R\times N_A\times N_E}$.

\subsubsection{Loss and class imbalance}

We used a weighted cross-entropy (wCE) \cite{he2009learning} plus soft dice (sDice) loss \cite{milletari2016v}:
\[
\mathcal{L} = \mathrm{CE}_w(Z,Y) + \lambda\,\mathrm{sDice}(Z,Y),\qquad \lambda=2.5,
\]
where $Y\in\{0,\ldots,K-1\}^{B\times N_R\times N_A\times N_E}$ are voxel labels. The class weights $w$ are calculated from the statistics of the data set and fixed during training (the released code includes exact values). 

\subsubsection{Training protocol}
Scenes 1, 3, 4, 5, 7 from the RaDelft dataset \cite{radelft_dataset} are used for training, and Scene 2 is used for testing.

\subsection{Semantic Fusion under Degraded Visual Conditions}
Adverse weather, such as fog, reduces the effectiveness of vision-based segmentation. However, radar is inherently robust to fog, making it a complementary sensing modality. To exploit this advantage, we fuse semantic information from camera and radar to get more reliable semantic priors for radar labeling under degraded visual conditions.
\subsubsection{Foggy image generation}
We generate fog images using fog simulation techniques for clear weather images in the RaDelft dataset \cite{radelft_dataset}. We adapt the pipeline used in \cite{FoggySynscapes} where $\beta$ is the attenuation coefficient of light. The larger the value of $\beta$, the denser the generated fog. In our experiment, we generate four different levels of fog, with the corresponding $\beta = 0.02, 0.04, 0.08, 0.15$.
\subsubsection{Radar depth-based segmentation}
We first employ the 4D radar point clouds from the RaDelft dataset and convert them into dense radar depth maps using the spectrum learning framework described in \cite{alsakabi2025toward}. This method learns to reconstruct the depth per pixel from the radar spectrum, yielding a radar depth map aligned with the camera view. We then train a DeepLabV3+ model \cite{chen2018encoder} in radar depth maps to obtain radar-based semantic segmentation results, which serve as a complement to the camera segmentation results.

\subsubsection{Camera-radar semantic fusion} \label{sec: semantic fusion}
\begin{figure}[htbp]
  \centering

  \begin{subfigure}{0.48\linewidth}
    \caption{Foggy image}
    \includegraphics[width=\linewidth]{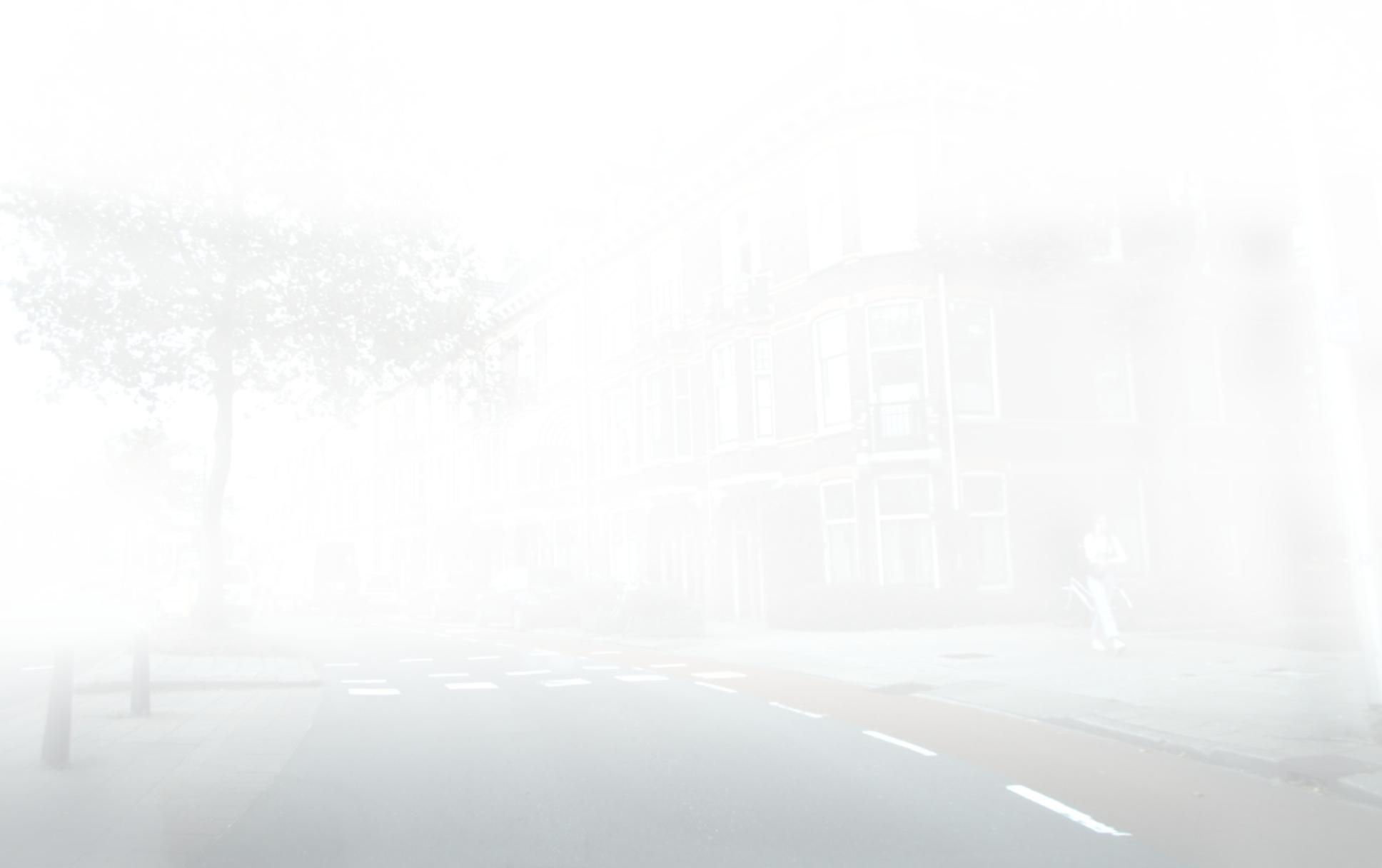}
  
  \end{subfigure}\hfill
  \begin{subfigure}{0.48\linewidth}
    \caption{Foggy image semantic segmentation}
    \includegraphics[width=\linewidth]{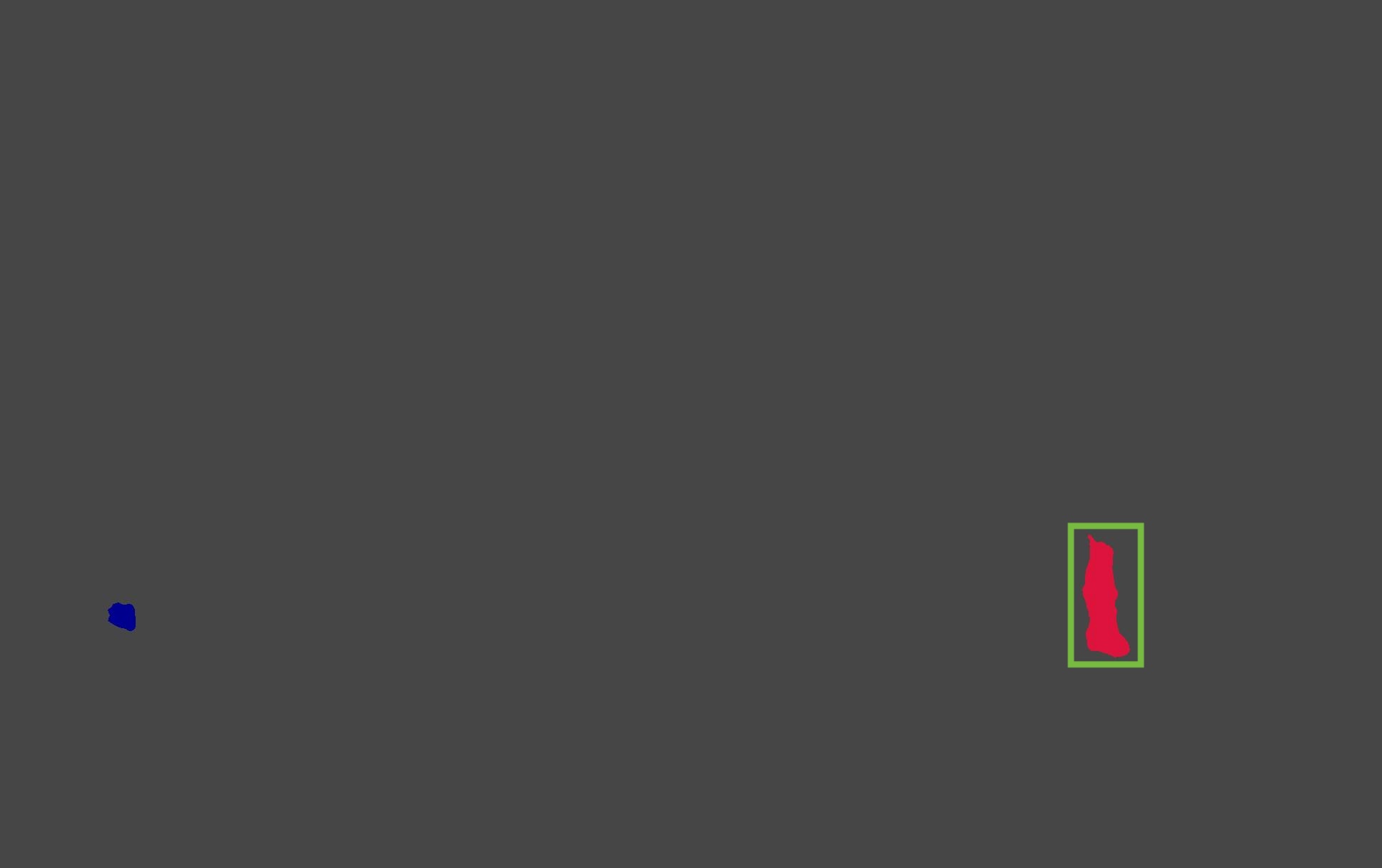}
   
  \end{subfigure}

  \vspace{0.4em}
  \begin{subfigure}{0.48\linewidth}
    \caption{Radar depth map}
    \includegraphics[width=\linewidth]{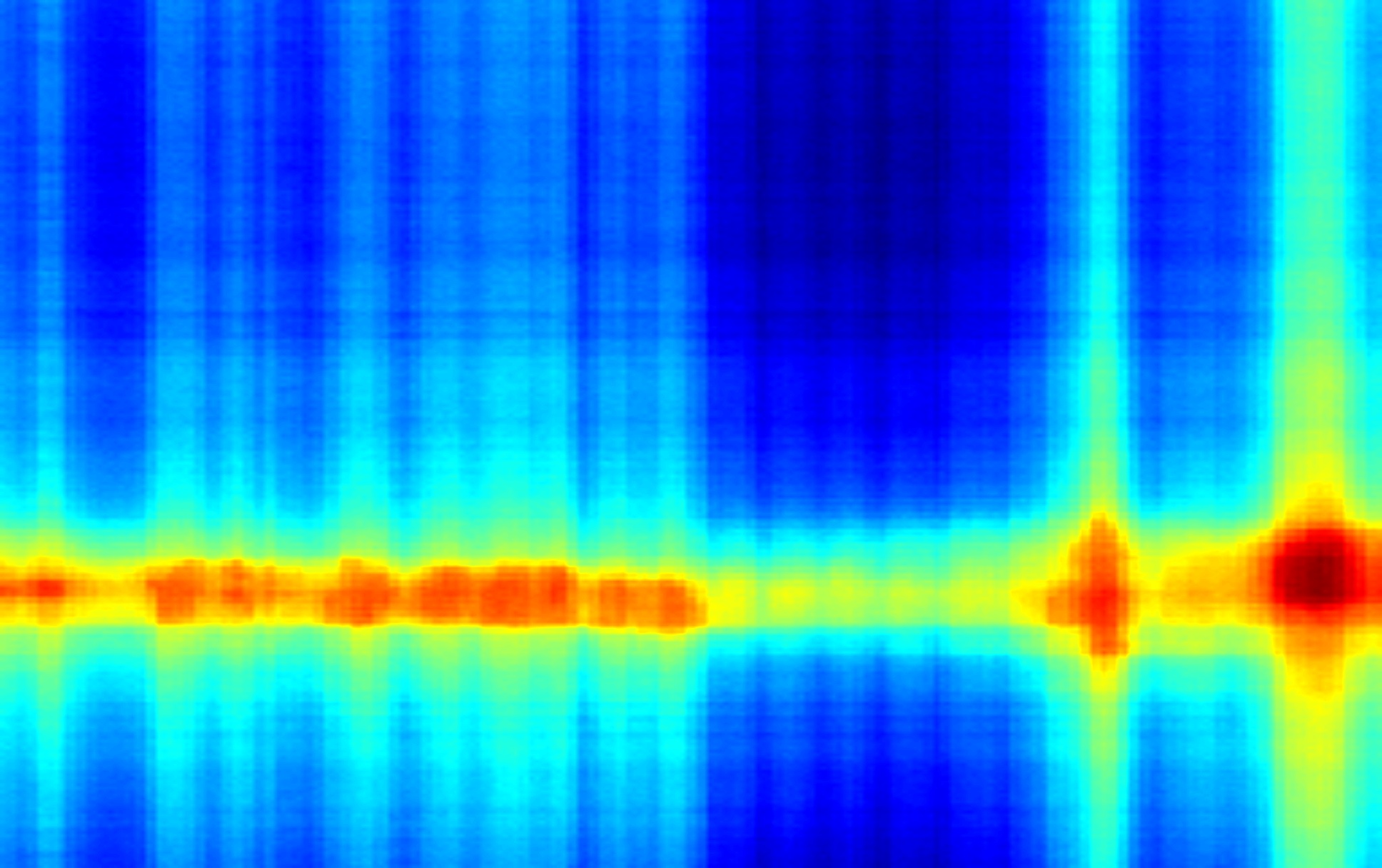}
  
  \end{subfigure}\hfill
  \begin{subfigure}{0.48\linewidth}
    \caption{Radar depth map semantic segmentation}
    \includegraphics[width=\linewidth]{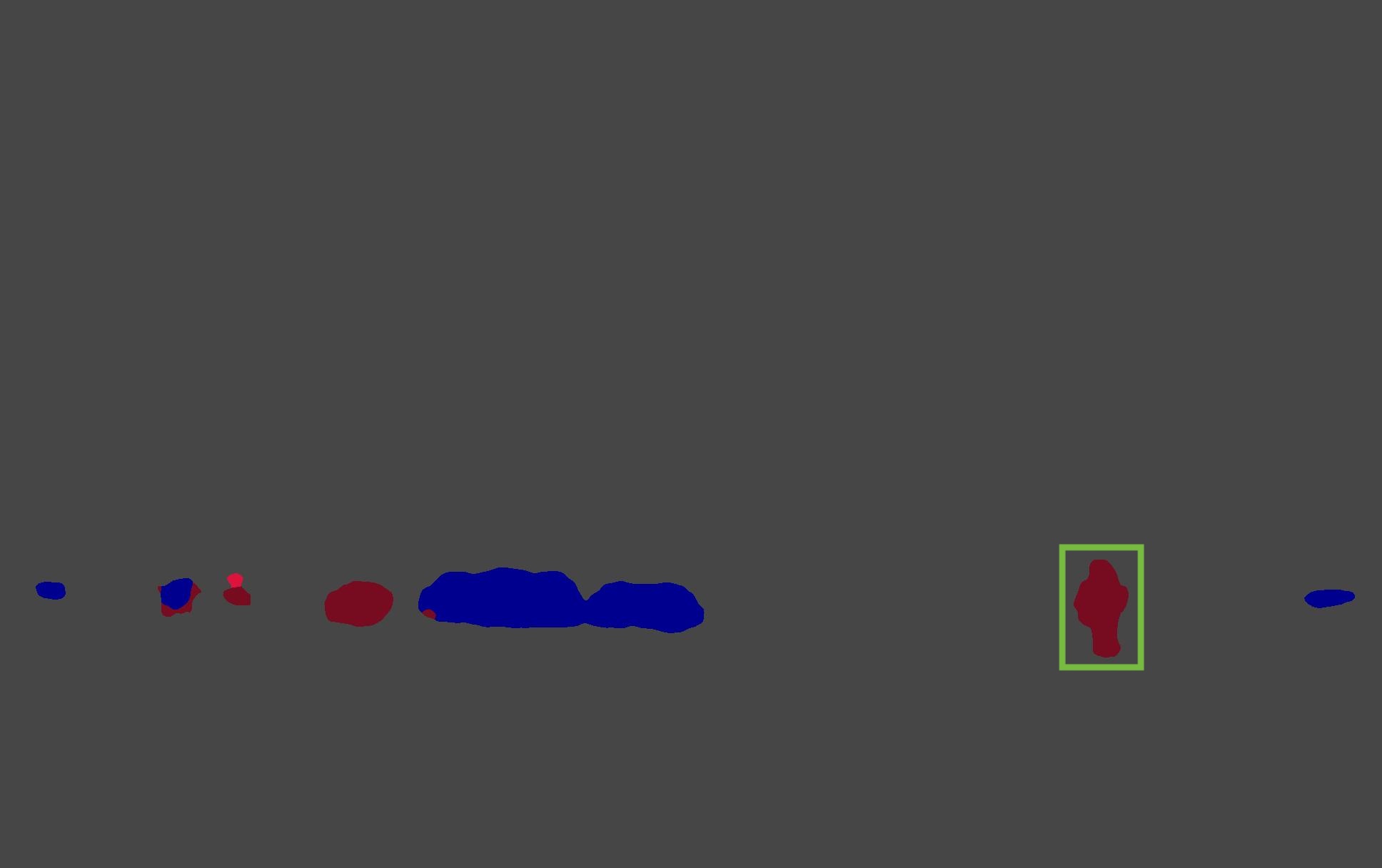}
   
  \end{subfigure}

  \vspace{0.4em}

  \begin{subfigure}{0.48\linewidth}
    \caption{Clear image}
    \includegraphics[width=\linewidth]{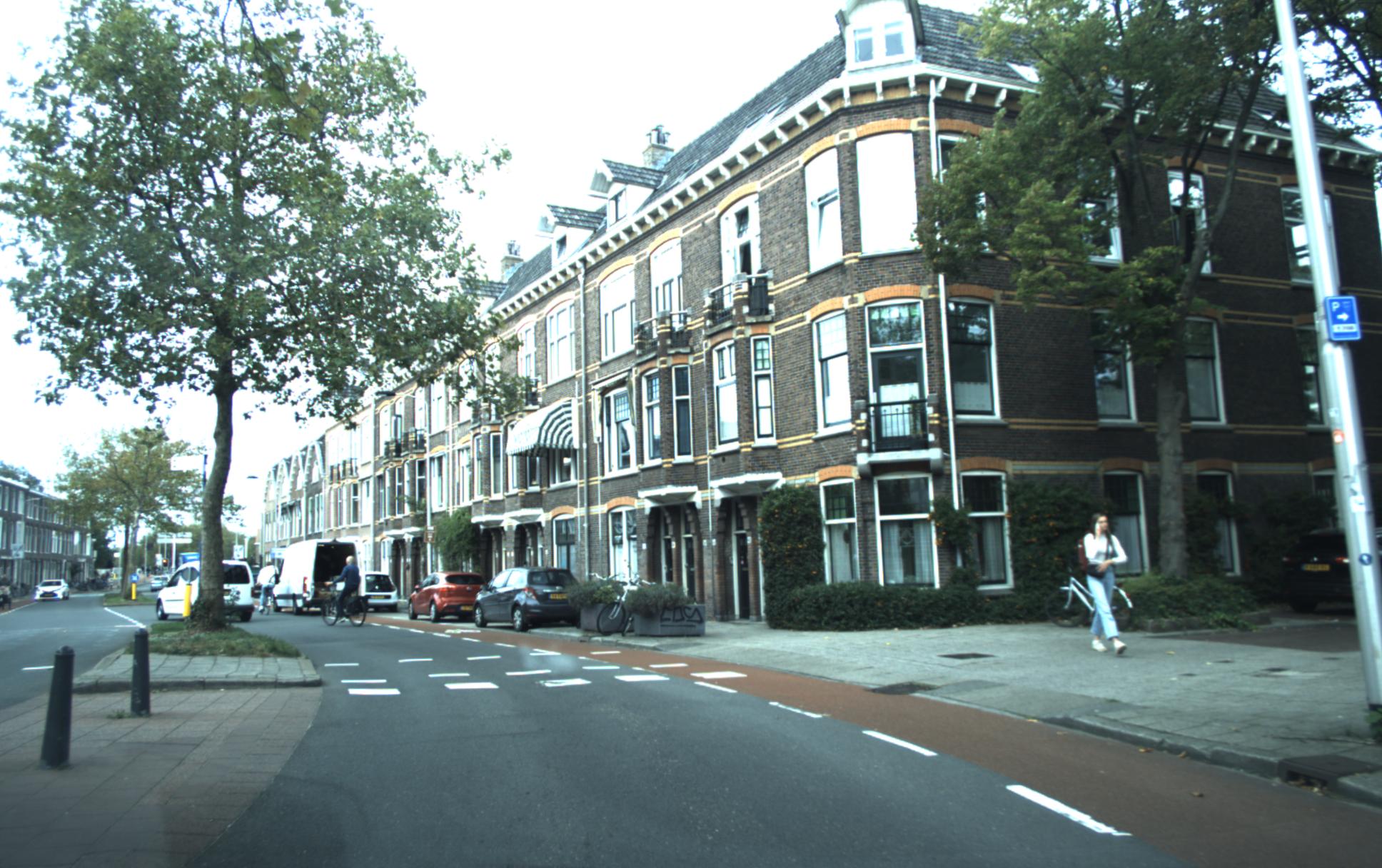}
  
  \end{subfigure}\hfill
  \begin{subfigure}{0.48\linewidth}
    \caption{Fused semantic segmentation}
    \includegraphics[width=\linewidth]{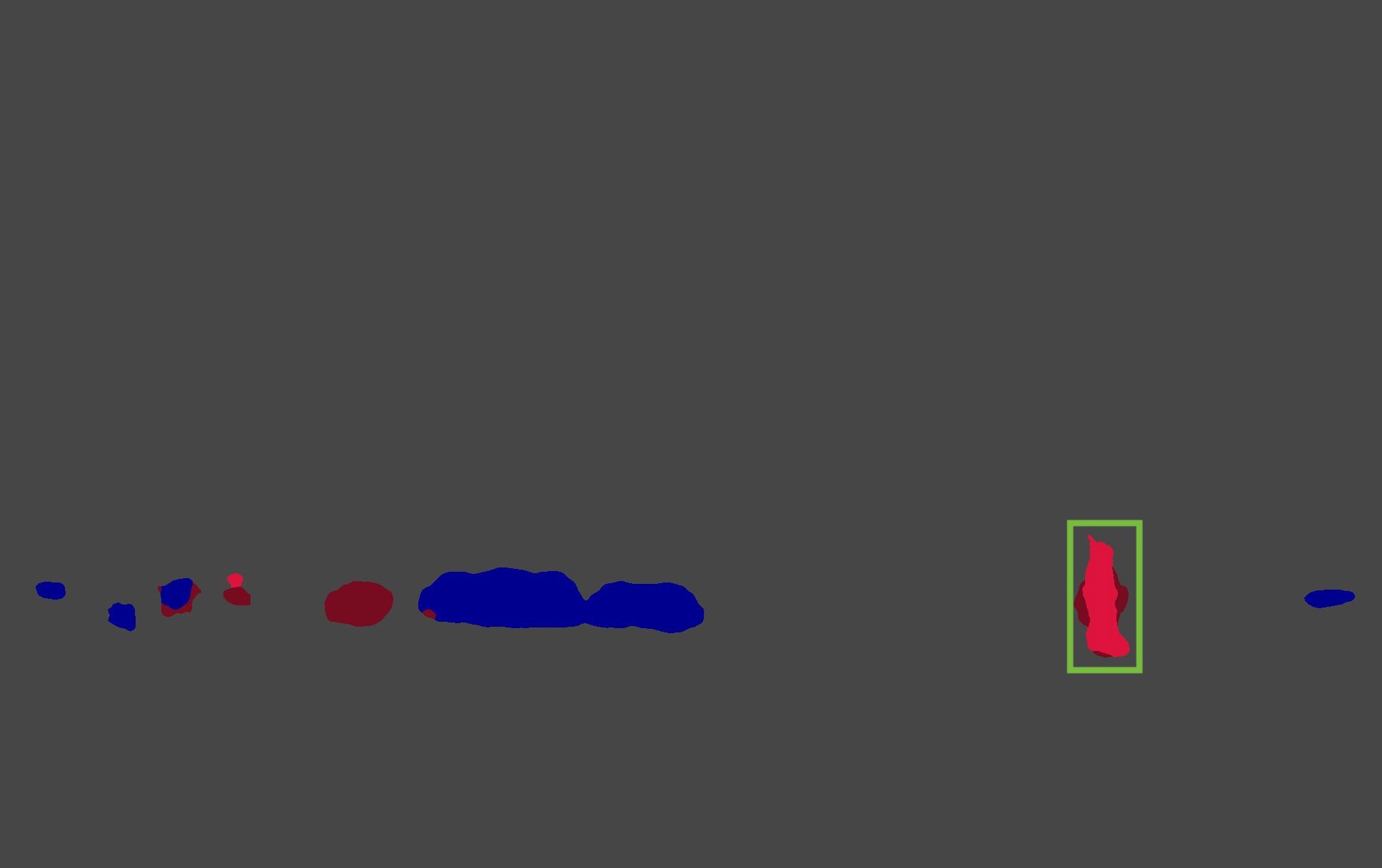}
   
  \end{subfigure}

  \caption{Example of semantic segmentation results. The first row shows the foggy image at fog intensity of $\beta = 0.15$ and its semantic segmentation. The second row shows the corresponding radar depth map and its segmentation result. The third row shows the clear image and the fused radar-camera segmentation. In semantic segmentation, dark grey represents scenario objects, pink represents pedestrians, blue represents vehicles and red represents bicycles. Plot (b) missed the vehicles due to the fog, but correctly identifies the person (bounded in green box highlighted in the figure). Plot (d) detected vehicles but misclassified the person as a cyclist. Plot (f) combines the vehicle detections from (d) with the correct person classification obtained in (b), showing a superior performance to camera-only (b) and radar-only (d).}
  \label{fig:seg_example}
\end{figure}

Given the semantic segmentation maps from the camera ($S_\text{cam}$) and radar depth maps ($S_\text{rad}$), we merge them into a unified semantic prior $S_\text{fuse}$ for labeling. Specifically, we treat the camera segmentation as the primary semantic source and use radar segmentation to complement missing targets when visual degradation causes the camera model to fail. The fusion logic is summarized as:
\[
S_\text{fuse}(p)=
\begin{cases}
S_\text{rad}(p), & \text{if } S_\text{cam}(p)=\text{background} \text{ and}\\
&  S_\text{rad}(p)\neq\text{background},\\[2pt]
S_\text{cam}(p), & \text{otherwise.}
\end{cases}
\]
Fig. \ref{fig:seg_example} shows examples of a camera segmentation result, radar segmentation result, and fused segmentation result.

\subsection{Camera-Assisted Labeling}
To have a better annotation for radar point clouds, we use a cross-modal geometric projection framework that transfers semantic information from camera images to radar point clouds. The complete pipeline is summarized below.
\subsubsection{Coordinate Alignment and Calibration}
Each radar frame is synchronized with the corresponding camera image through the dataset timestamps. The intrinsic and extrinsic parameters are provided from the dataset. Given the roll-pitch-yaw angles $(r, p, y)$ and translation vector $\mathbf{t}$, a homogeneous transformation matrix is constructed to map radar-frame coordinates into the camera coordinate system:
\begin{equation}
    T_{\text{cam} \leftarrow \text{rad}} = \begin{bmatrix}
        R_z(y) \, R_y(p) \, R_x(r) & \mathbf{t} \\
        \mathbf{0}^\top & 1
    \end{bmatrix}
\label{eq:transform_matrix}
\end{equation}
where $R_x(r)$, $R_y(p)$, and $R_z(y)$ denote the basic rotation matrices around the $x$, $y$, and $z$ axes. The radar point cloud $\mathbf{P} = [x, \, y, \, z, 1]^\top$ is transformed into the camera coordinate frame as
\begin{equation}
    \mathbf{P}_{\text{cam}} = T_{\text{cam} \leftarrow \text{rad}} \, \mathbf{P}
    \label{eq:transform_points}
\end{equation}

\subsubsection{3-D $\rightarrow$ 2-D Projection}
After transforming the radar point cloud into the camera coordinate frame, each point $\mathbf{P}_{\text{cam}} = [X_x, \, Y_c, \, Z_c, \, 1]^\top$ is projected onto the camera image plane using the camera projection matrix~$K$.
The pixel coordinates $(u, v)$ are obtained by the perspective projection and homogeneous normalization:
\begin{equation}
    \tilde{\mathbf{u}} = K \, \mathbf{P}_{\text{cam}}
    \label{eq:proj_homo}
\end{equation}
\begin{equation}
    u = \frac{\tilde{u}}{\tilde{w}}, \qquad
    v = \frac{\tilde{v}}{\tilde{w}}, \qquad
    d = Z_c,
    \label{eq:pixel_coords}
\end{equation}
where $d$ represents the depth of the radar return along the camera range axis.
Only points that fall within the valid image region and have positive depth are retained. Let the image resolution be $(H, \, W)$; then the visibility mask is defined as 
\begin{equation}
    \mathcal{M} = \{\,i \mid 0 < u_i < W,\; 0 < v_i < H,\; d_i > 0 \,\}
    \label{eq:mask_valid}
\end{equation}
In practice, we also reject distant points whose range exceeds a predefined threshold ($d_i < 50$\,m in our implementation) to avoid extremely sparse radar reflections. 
\subsubsection{Semantic Transfer} \label{sec: semantic_transfer}
For each valid projected point $(u, v)$, the semantic class label is sampled from the corresponding pixel of the camera semantic segmentation map produced by a pre-trained DeepLabV3 + model \cite{chen2018encoder}.
Each valid radar point $(x_i, y_i, z_i)$ is associated with its projected pixel location $(u_i, v_i)$, allowing pixel-level semantic sampling from the camera segmentation map.
\begin{equation}
c_i = S_{\text{img}}(u_i, v_i), \qquad i \in \mathcal{M},
\label{eq:sample_semantics}
\end{equation}
where $S_{\text{img}}$ denotes the semantic segmentation function of the camera that returns a class index for pixel $(u, v)$. This establishes a one-to-one correspondence between visible radar returns and image pixels, which forms the basis for cross-modal semantic label transfer.
\subsubsection{Post-Processing and Refinement} \label{sec: refinement}
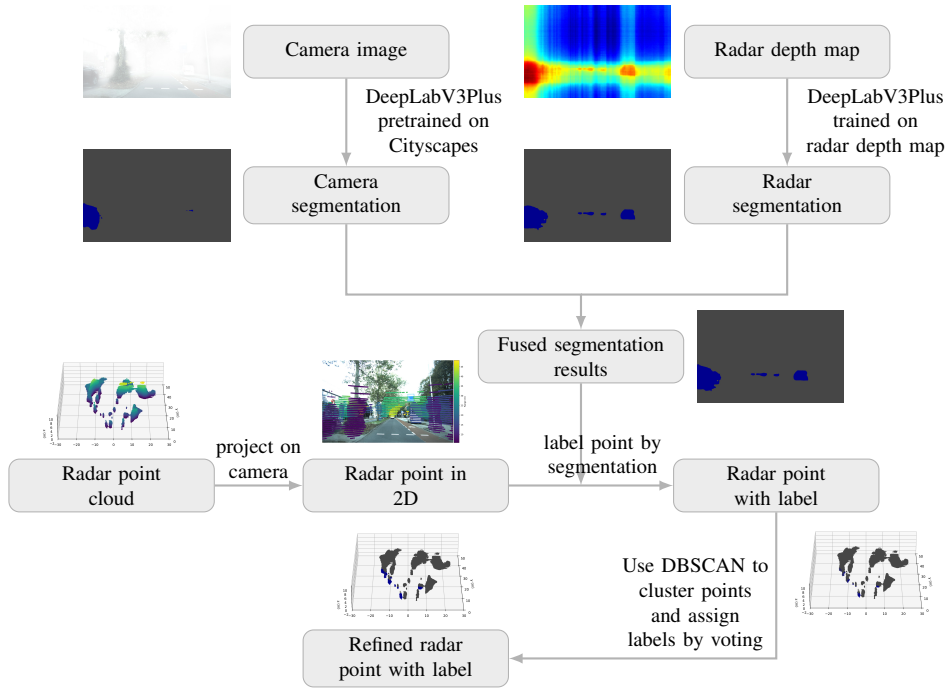
\begin{figure*}[t]
    \centering
    \begin{tikzpicture}[
        scale=0.78,
        every node/.style={transform shape},
        >=latex,
        node distance = 8mm and 15mm,
        line/.style={thick, draw=gray!60},
        block/.style={
            draw=gray!60,
            rounded corners,
            fill=gray!15,
            minimum width=35mm,
            minimum height=9mm,
            align=center
        },
        note/.style={
            align=center,
        },
        textnode/.style={
            align=center
        }
    ]
    
    
    \node[block] (camimg) {Camera image};
    \node[left=1mm of camimg] (camimg_pic) {\includegraphics[width=2.5cm]{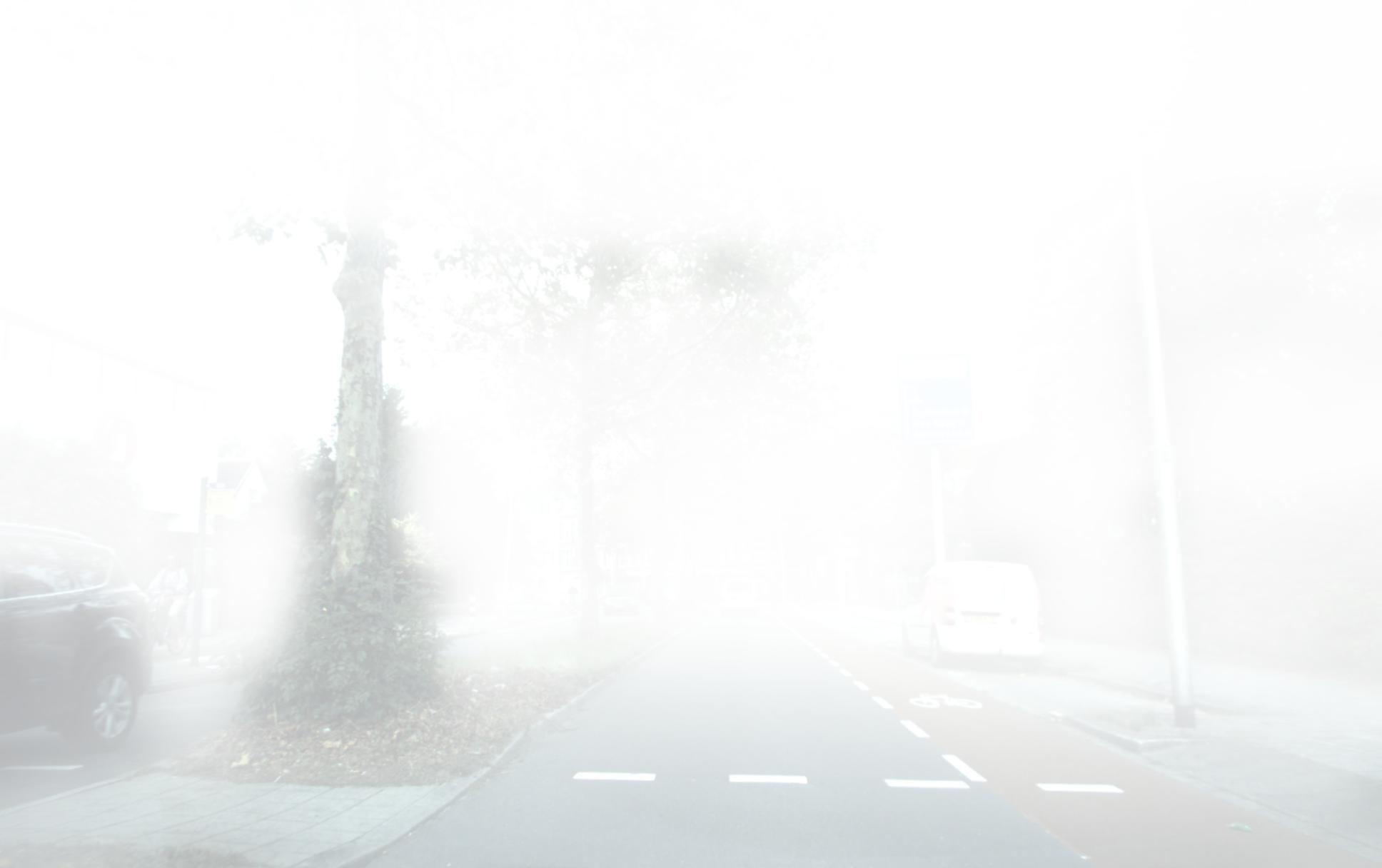}};
    \node[block, right=40mm of camimg] (radarimg) {Radar depth map};
    \node[left=1mm of radarimg] (radarimg_pic) {\includegraphics[width=2.5cm]{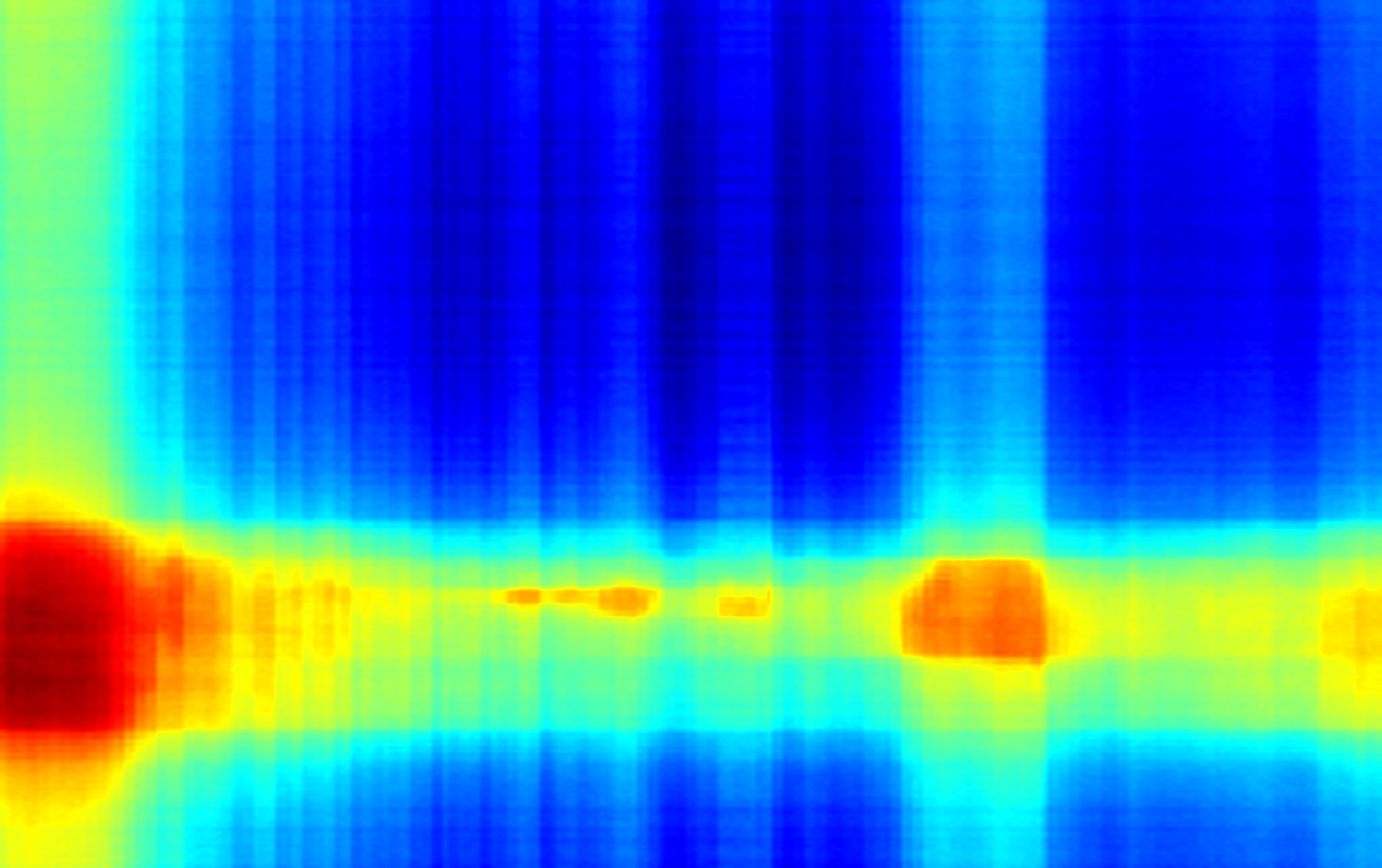}};
    
    
    \node[block, below=15mm of camimg] (camseg) {Camera\\segmentation};
    \node[left=1mm of camseg] (camseg_pic) {\includegraphics[width=2.5cm]{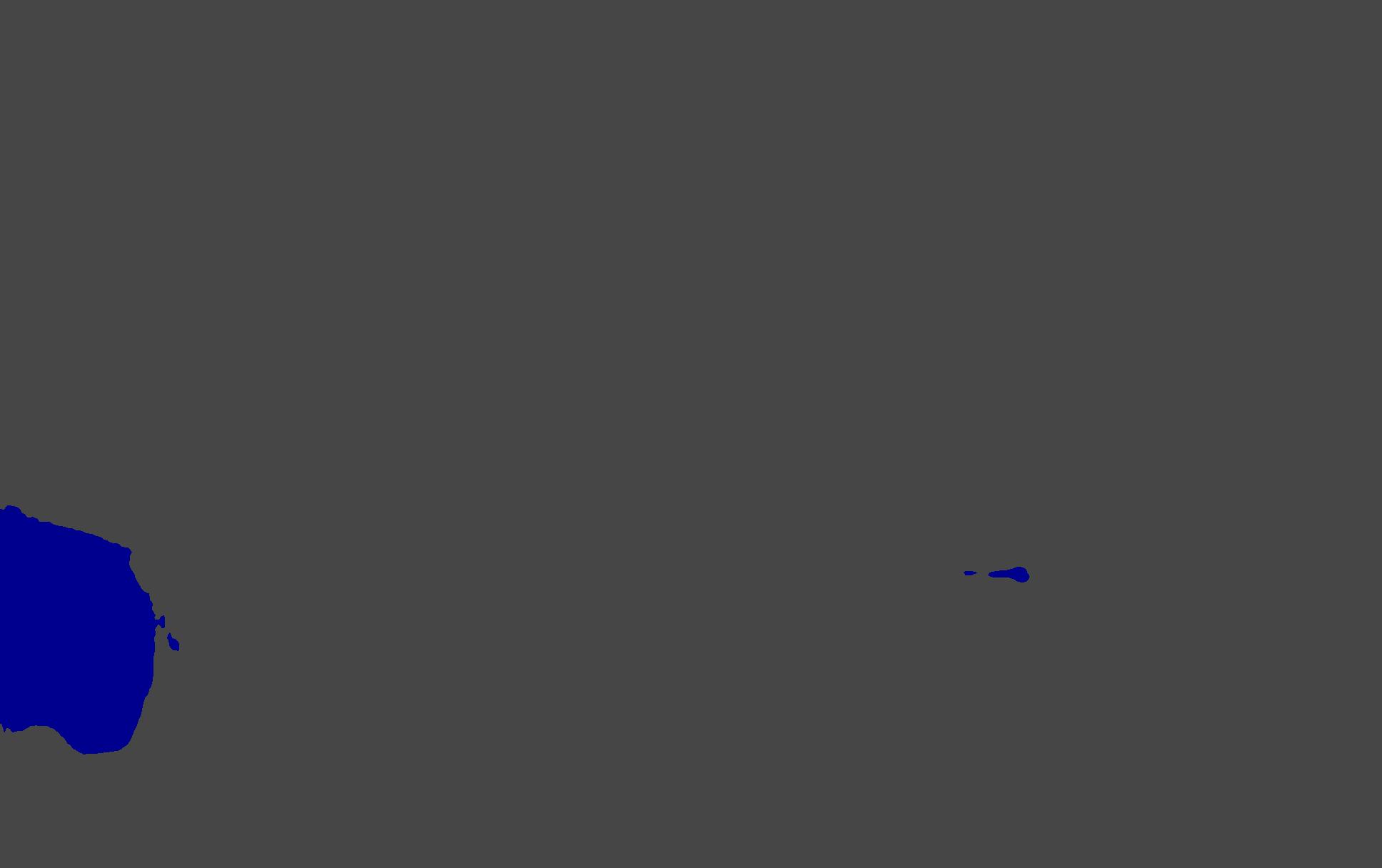}};
    \node[block, below=15mm of radarimg] (radarseg) {Radar\\segmentation};
    \node[left=1mm of radarseg] (radarseg_img) {\includegraphics[width=2.5cm]{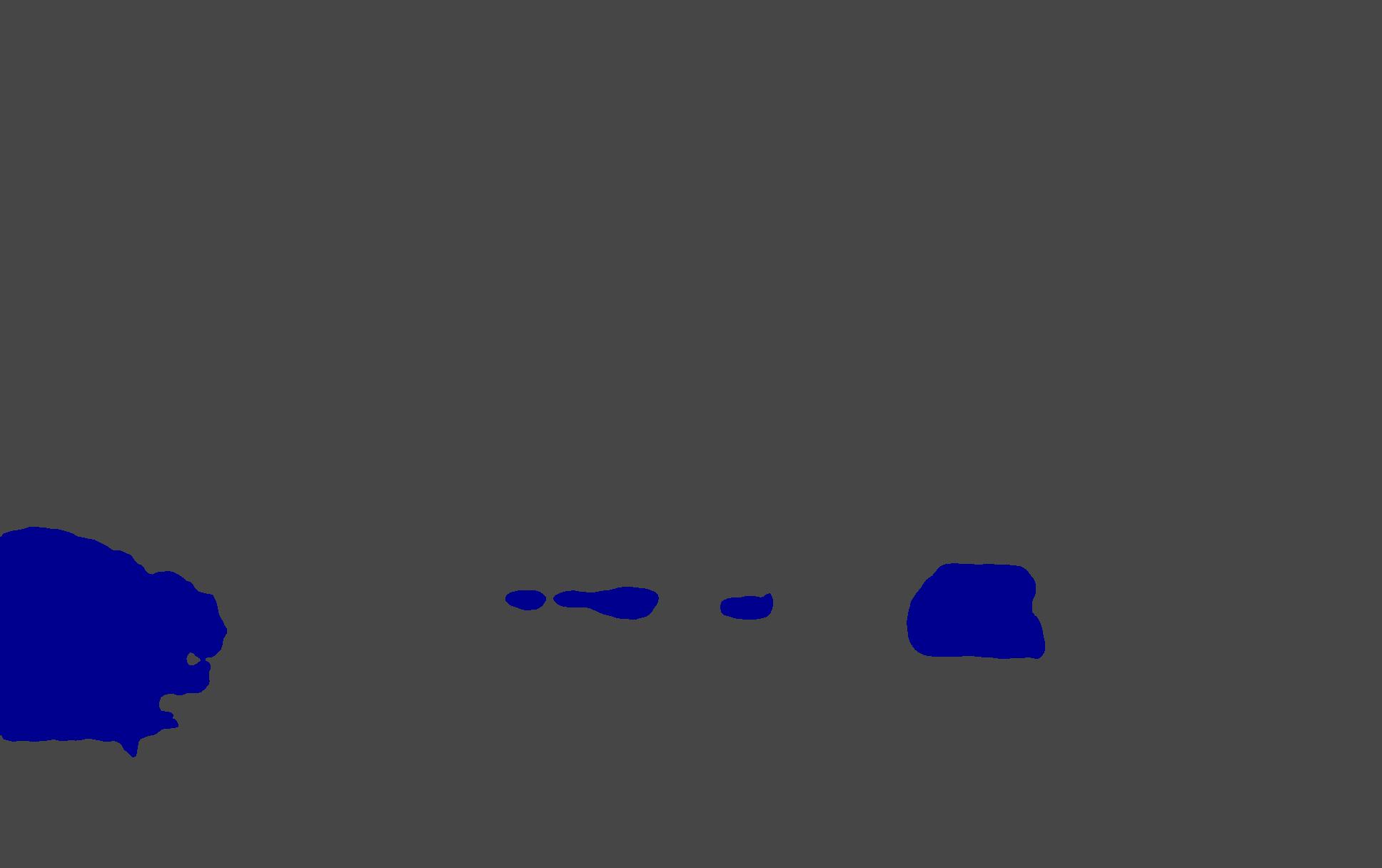}};
    
    \draw[line,->] (camimg.south) -- node[right, note, xshift=2mm]
        {DeepLabV3Plus\\pretrained on\\Cityscapes}
        (camseg.north);
    
    \draw[line,->] (radarimg.south) -- node[right, note, xshift=2mm]
        {DeepLabV3Plus\\trained on\\radar depth map}
        (radarseg.north);
    
    
    \coordinate[below=12mm of camseg, xshift=40mm] (mergepoint) {};
    
    \node[block, below=6mm of mergepoint] (fused)
        {Fused segmentation\\results};
    \node[right=1mm of fused] (fused_pic) {\includegraphics[width=2.5cm]{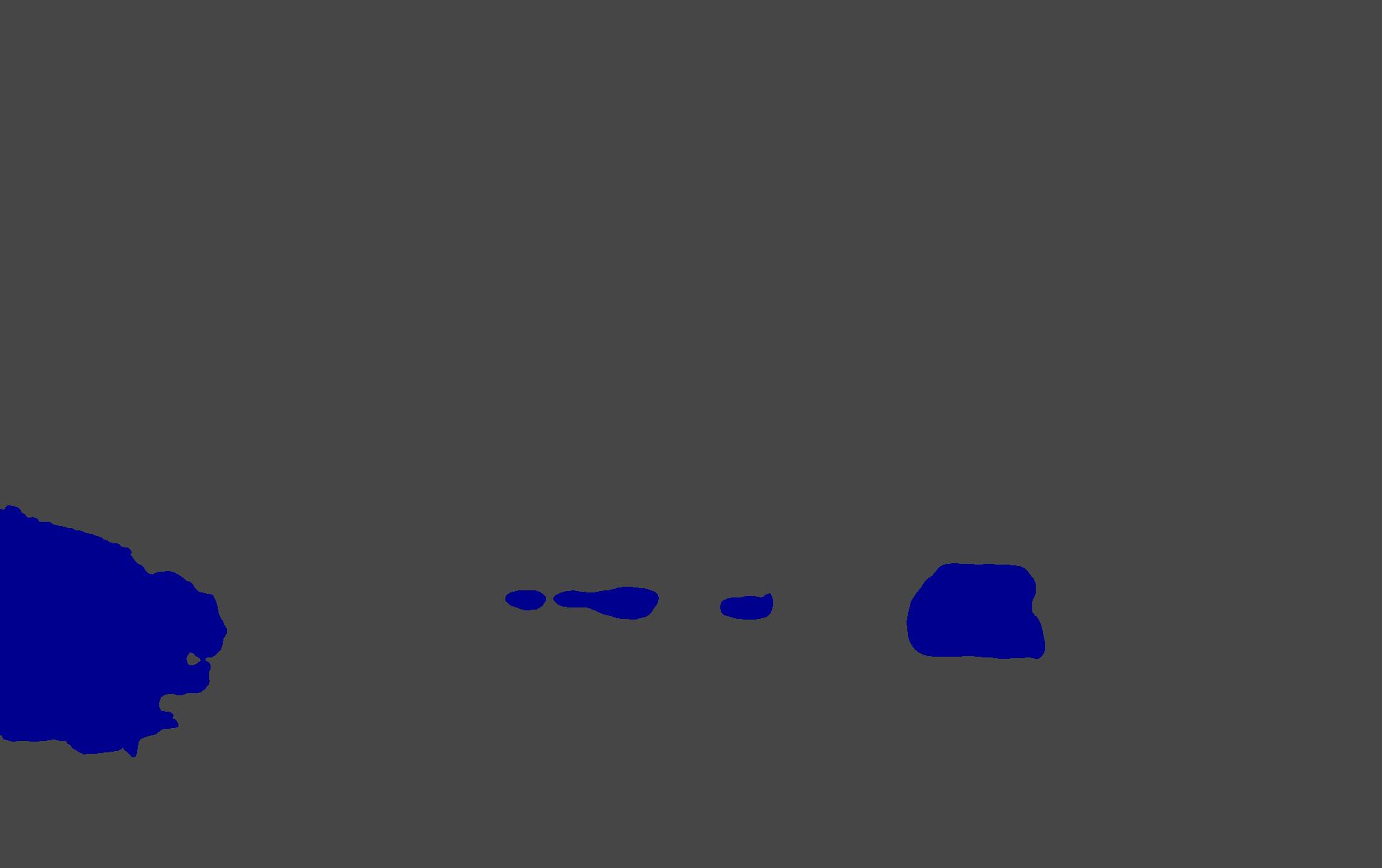}};
    
    \draw[line,->] (camseg.south) |- (mergepoint) -- (fused);
    \draw[line] (radarseg.south) |- (mergepoint);
    
    
    \node[block, below=40mm of camseg, xshift=-40mm] (radarpc)
        {Radar point\\cloud};
    \node[above=1mm of radarpc] (radarpc_pic) {\includegraphics[width=2.5cm, trim = 0mm 30mm 0mm 30mm, clip]{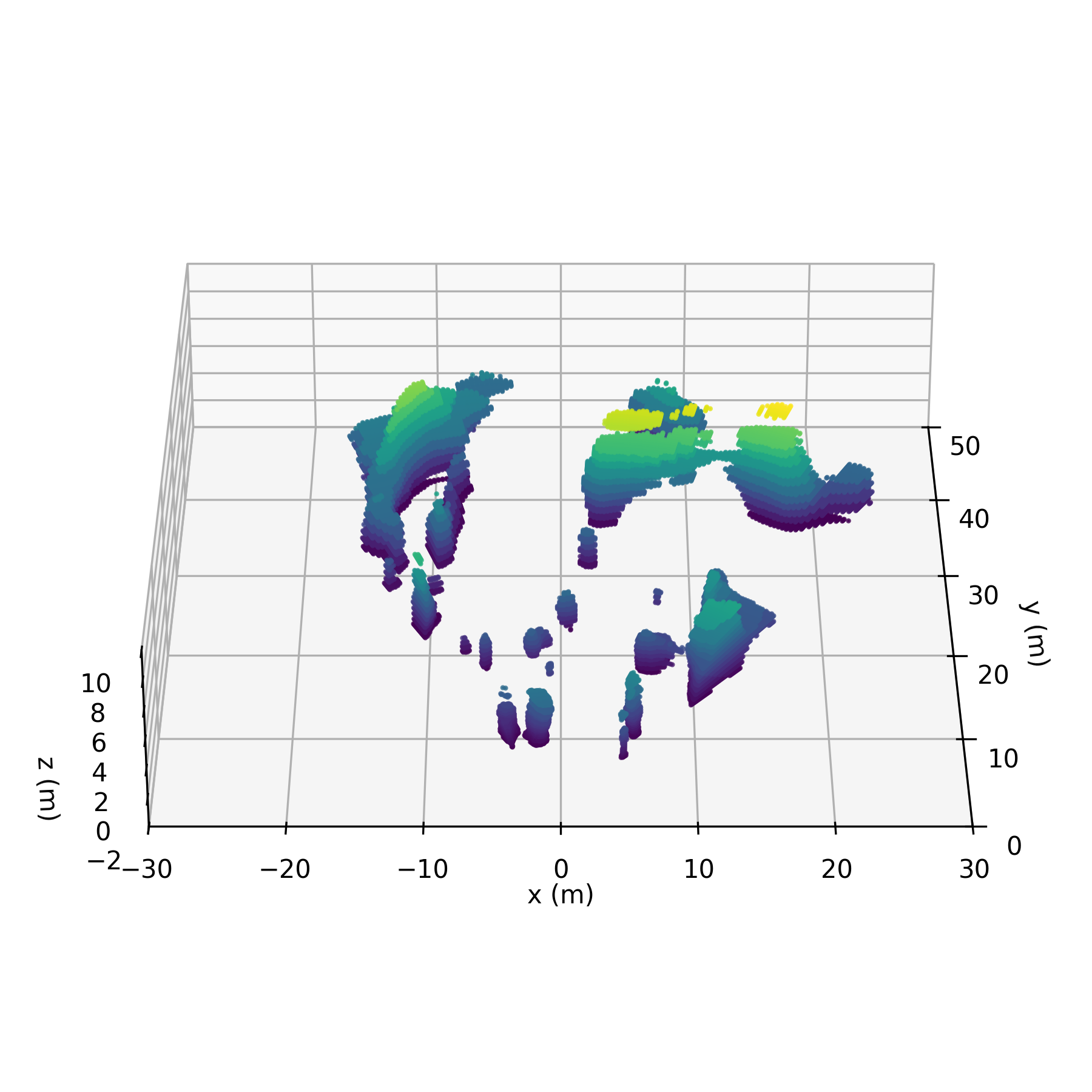}};
    
    \node[block, right=15mm of radarpc] (radar2d)
        {Radar point in\\2D};
    \node[above=1mm of radar2d, xshift=-2mm] (radar2d_pic) {\includegraphics[width=2.5cm]{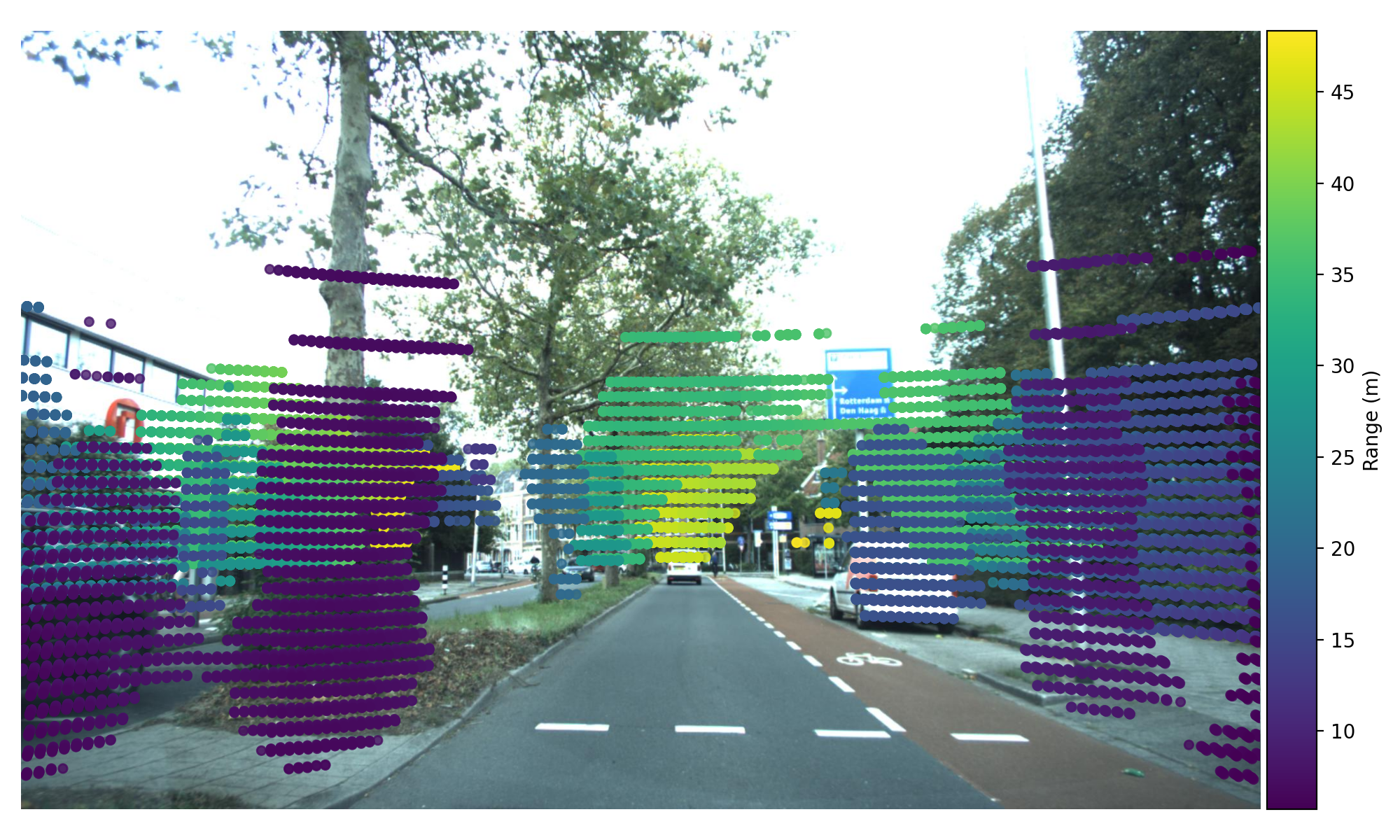}};
        
    \node[block, right=28mm of radar2d] (radarlabel)
        {Radar point\\with label};
    \node[below=1mm of radarlabel, xshift=1.5cm] (radarlabel_pic) {\includegraphics[width=2.5cm, trim = 0mm 30mm 0mm 30mm, clip]{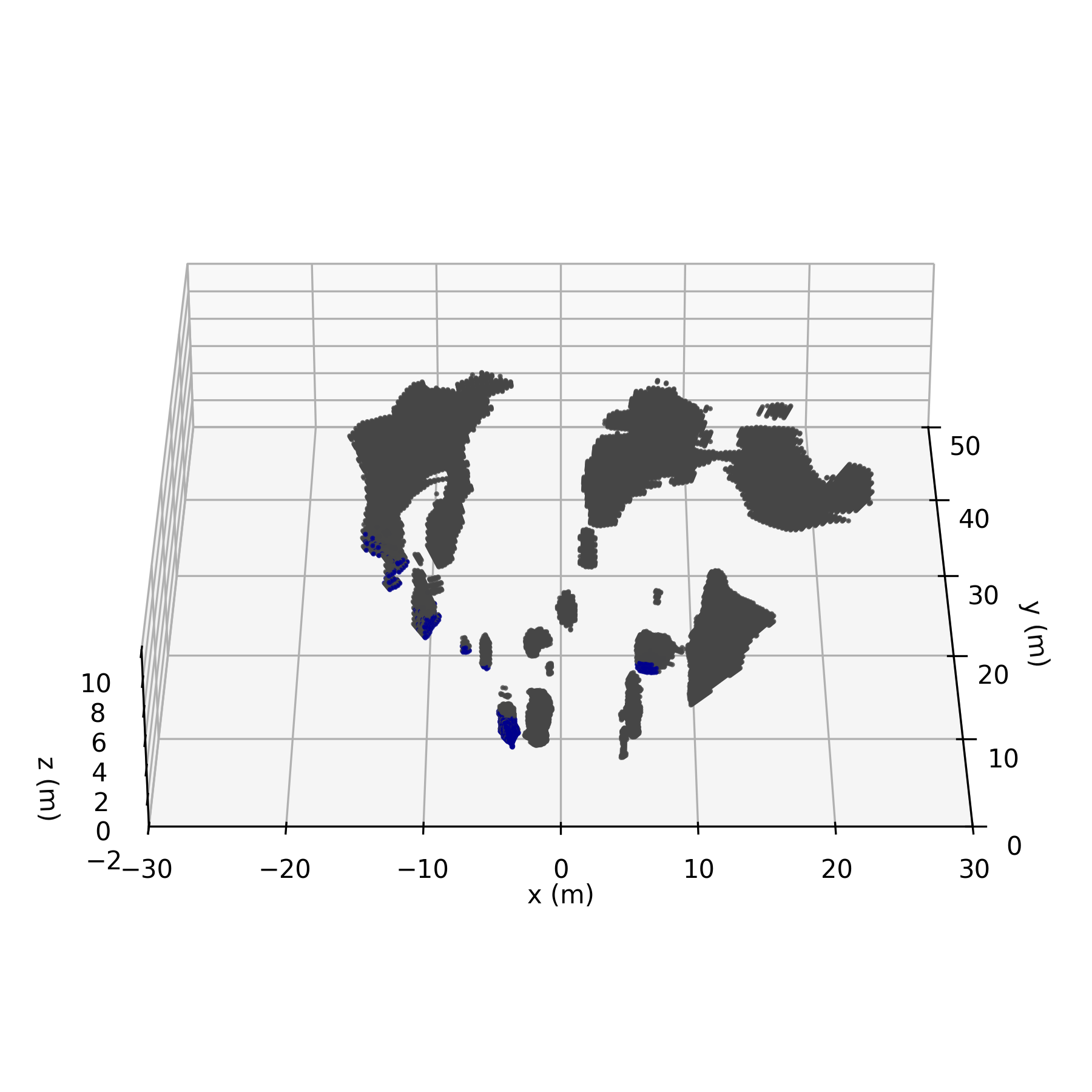}};
    
    \draw[line,->] (fused) -- (fused|-radarlabel);
    \draw[line,->]
        (radarpc.east) -- node[above, textnode]{project on\\camera}
        (radar2d.west);
    

    \draw[line,->]
        (radar2d.east) -- node[above, textnode,xshift=2mm]{label point by\\ segmentation}
        (radarlabel.west);
    
    
    \node[block, below=20mm of radar2d] (refined)
        {Refined radar\\point with label};
    \node[above=1mm of refined] (refined_pic) {\includegraphics[width=2.5cm, trim = 0mm 30mm 0mm 30mm, clip]{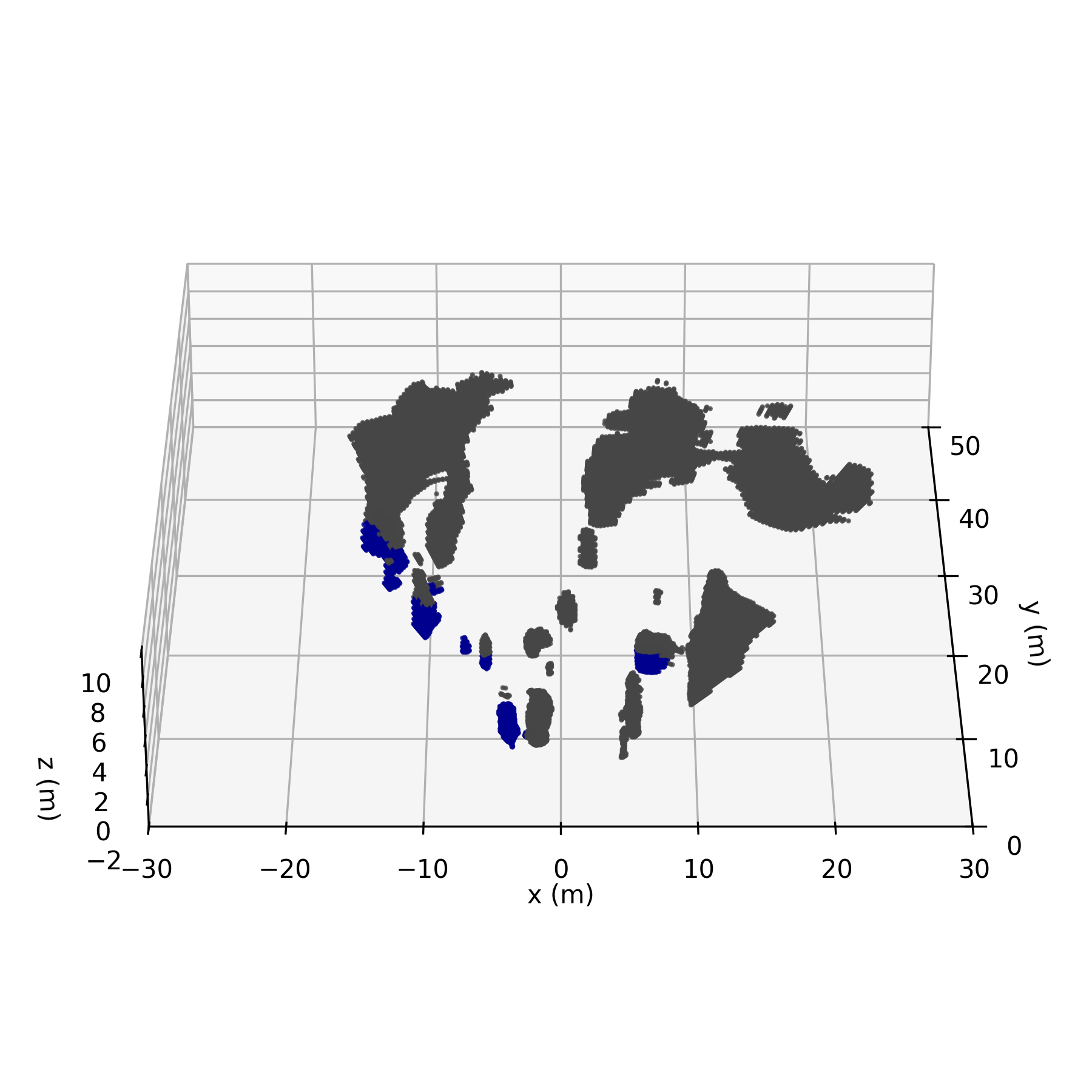}};
        
    \draw[line,->]
        (radarlabel.south) |- node[above left, textnode]
            {Use DBSCAN to\\cluster points\\and assign\\labels by voting}
        (refined.east);
    
    \end{tikzpicture}

    \caption{Radar–camera fusion and radar point cloud labeling pipeline.}
    \label{fig:fusion_pipeline}
\end{figure*}

Although projected camera labels provide dense supervision, misalignments and noise remain due to occlusion and calibration errors. To improve radar labeling, we refined the labeled radar points using a cluster-based filtering scheme. First, we group radar points with density-based spatial clustering of applications with noise (DBSCAN) in 3-D space to form spatial clusters. Within each cluster, the proportion of object labels (pedestrian, vehicle, bicycle) is compared with class-specific thresholds. If only one class exceeds its threshold, the entire cluster is reassigned to that class; otherwise it reverts to the background. Finally, to prevent isolated false positives, we validate the clusters by nearest-neighbor distance. Only points within a class-dependent radius are retained. This refinement step gives labels to points that are occluded by foreground points and ensures local spatial coherence.
\newline
\newline
Fig. \ref{fig:fusion_pipeline} illustrates the camera-radar fusion labeling pipeline. Camera image and radar depth maps are segmented separately, and fused into a joint semantic map as described in Sec \ref{sec: semantic fusion}. This map is used to assign per-point labels after projecting the radar point cloud into the camera frame, as described in Sec \ref{sec: semantic_transfer}. The resulting labeled radar points are further spatially refined using the method stated in Sec \ref{sec: refinement}.

\section{Experiments}
\subsection{Reproducing the RaDelft Baseline}
We re-implemented the radar semantic segmentation pipeline described in the RaDelft work. We train for up to 60 epochs on scenes \{1, 3, 4, 5, 7\}, with an 80/20 train/val split. Scene~2 is held out for testing. The loss function is a weighted combination of standard cross-entropy and multi-class soft Dice loss. The cross-entropy term uses per-class weights to counter the extreme class imbalance between classes (over $99\%$ of the voxels are empty, and among those voxels that are not empty, $92\%$ belong to scenario objects). We observed that these class weights are critical to the final performance: small changes in the weights can significantly shift the trade-off between high $P_d$ (detection rate) and low $P_{fa}$ (false alarm rate). In practice, we swept multiple weight configurations for the five classes and selected a setting that is closest to the original paper. The per-class weights we used is 
\[
\mathbf{w} = [1.27\times10^{-4},\; 2.26\times10^{-2},\; 5.99,\; 3.93\times10^{-1},\; 2.50],
\]
where the weights correspond respectively to the classes empty, scenario objects, pedestrians, vehicles, and bicycles. This tuning is necessary because the radar voxel distribution is highly skewed: most voxels are empty, while pedestrians and cyclists occupy only a tiny fraction of the volume. Without reweighting, the model collapses toward predicting mostly empty and achieves artificially low $P_{fa}$ but very poor object $P_d$. With stronger weights on minority classes, the network becomes more willing to detect small targets, which increases $P_d$ for vehicles and VRUs (‘pedestrians’ and ‘bicycles’ are combined in the same class to consider vulnerable road users) but can also raise $P_{fa}$ if the weights are too aggressive.

\subsection{Radar Depth Map Semantic Segmentation}
To obtain radar-based semantic information, we trained a DeepLabV3+ network with a ResNet-101 backbone on the radar depth maps derived from the RaDelft dataset. The ground-truth labels for training the radar depth map segmentation network were obtained from the semantic segmentation results of the corresponding camera images. A DeepLabV3+ model pretrained on the Cityscapes dataset was used to generate pixel-level semantic maps for each image. The resulting segmentation masks were then converted into five unified categories (background, scenario objects, pedestrians, vehicles, and bicycles). Each radar frame was converted into a pseudo-image representation \cite{alsakabi2025toward} and labeled according to the same five semantic classes. We generate $1500$ frames of radar depth map: the first $150$ frames are left for testing, the rest are used for training.
\section{Evaluation and Results}
\subsection{Reproduce Performance}
To evaluate the reproducibility performance, we compare our reported results with the original RaDelft baseline reported by Sun~\cite{sun2025automatic}. All experiments are conducted on the same dataset configuration and use identical radar cube dimensions and evaluation metrics. Table \ref{tab:reproduce} presents the detection probability ($P_\mathrm{d}$), false alarm probability ($P_\mathrm{fa}$), and chamfer distance (CD) for different object categories. As shown in Table \ref{tab:reproduce}, our reproduced results closely match the quantitative performance reported in the original RaDelft thesis \cite{sun2025automatic} in all metrics. Fig. \ref{fig:reproduce} shows an example of the generated
radar point cloud (PC) with class information, the corresponding LiDAR
PCs ground truth, and the reference camera image.

\begin{table*}[ht] 
\centering
\caption{Quantitative comparison between the RaDelft baseline and our reproduced model.}
\label{tab:reproduce}
\renewcommand{\arraystretch}{1.15}
\setlength{\tabcolsep}{4pt} 
\begin{tabular}{l c c c c c c c c c c c}
\toprule
\textbf{Method} &
$P_{d\_\mathrm{All}}$ &
$P_{fa\_\mathrm{All}}$ &
$P_{d\_\mathrm{Scenario}}$ &
$P_{fa\_\mathrm{Scenario}}$ &
$P_{d\_\mathrm{Vehicles}}$ &
$P_{fa\_\mathrm{Vehicles}}$ &
$P_{d\_\mathrm{VRU}}$ &
$P_{fa\_\mathrm{VRU}}$ &
$CD_\mathrm{All}$ &
$CD_\mathrm{Scenario}$ &
$CD_\mathrm{Target}$ \\
\midrule
Delft &
63.1\% & 3.55\% & 60.7\% & --- &
34.8\% & --- &
23.1\% & --- &
2.15\,m & 2.45\,m & 6.73\,m \\
Ours &
65.1\% & 2.23\% &
63.2\% & 2.09\% &
42.3\% & 0.11\% &
28.9\% & 0.04\% &
1.72\,m & 1.96\,m & 6.53\,m \\
\bottomrule

\end{tabular}
\\[2pt]
\raggedright
\footnotesize Note: --- indicates a value not reported by the original author.
\end{table*}

\definecolor{ScenarioGray}{RGB}{70,70,70}
\definecolor{PedRed}{RGB}{220,20,60}
\definecolor{VehBlue}{RGB}{0,0,142}
\definecolor{BikeMaroon}{RGB}{119,11,32}

\captionsetup[subfigure]{position=top, justification=centering, skip=2pt, labelformat=empty}

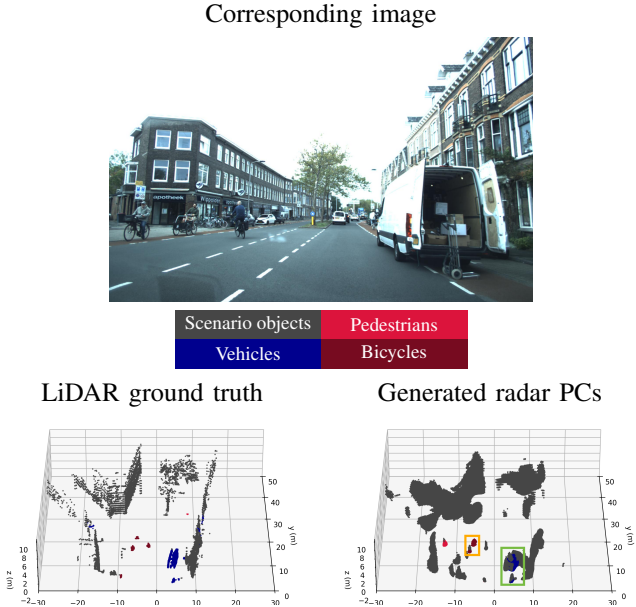
\begin{figure}[htbp]
  \centering

  \begin{subfigure}{0.65\linewidth}
    \caption{Corresponding image} 
    \includegraphics[width=\linewidth]{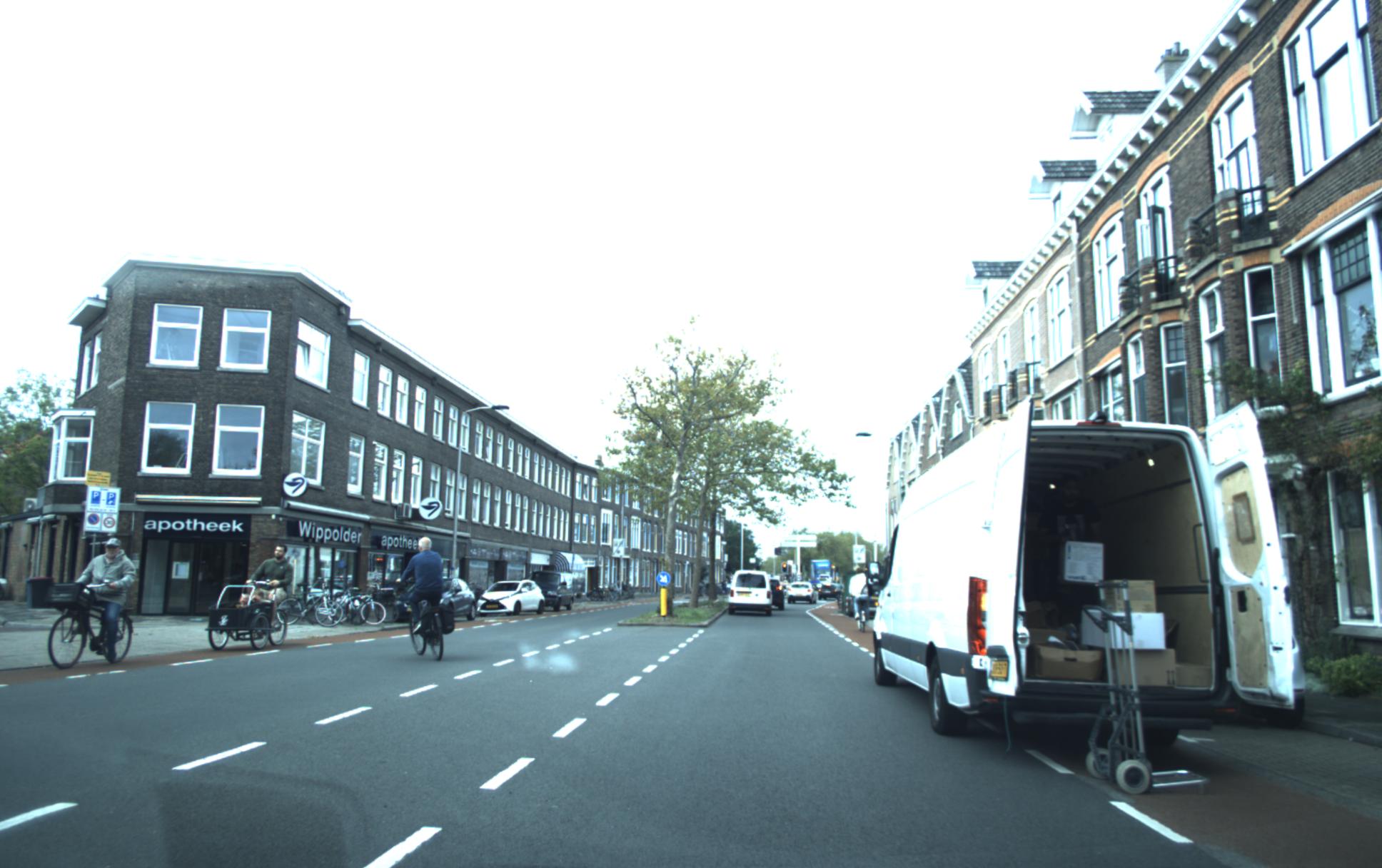}
  
  \end{subfigure}

  \vspace{0.3em}

  \begin{subfigure}{0.45\linewidth}
  \centering
  \resizebox{\linewidth}{!}{%
  \begin{tikzpicture}[x=1cm,y=1cm,baseline]
    \def\W{7.5} \def\H{1.5}
    \def\cw{0.5*\W} \def\ch{0.5*\H}

    \fill[ScenarioGray] (0,\ch) rectangle (\cw,\H);
    \fill[PedRed]       (\cw,\ch) rectangle (\W,\H);
    \fill[VehBlue]      (0,0)     rectangle (\cw,\ch);
    \fill[BikeMaroon]   (\cw,0)   rectangle (\W,\ch);

    \node[white,font=\Large] at (0.5*\cw, \ch+0.5*\ch) {Scenario objects};
    \node[white,font=\Large] at (\cw+0.5*\cw, \ch+0.5*\ch) {Pedestrians};
    \node[white,font=\Large] at (0.5*\cw, 0.5*\ch) {Vehicles};
    \node[white,font=\Large] at (\cw+0.5*\cw, 0.5*\ch) {Bicycles};
  \end{tikzpicture}}

  \label{fig:legend}
  \end{subfigure}

  \vspace{0.4em}

  \begin{subfigure}{0.48\linewidth}
    \caption{LiDAR ground truth}
    \includegraphics[width=\linewidth, trim = 0mm 30mm 0mm 30mm, clip]{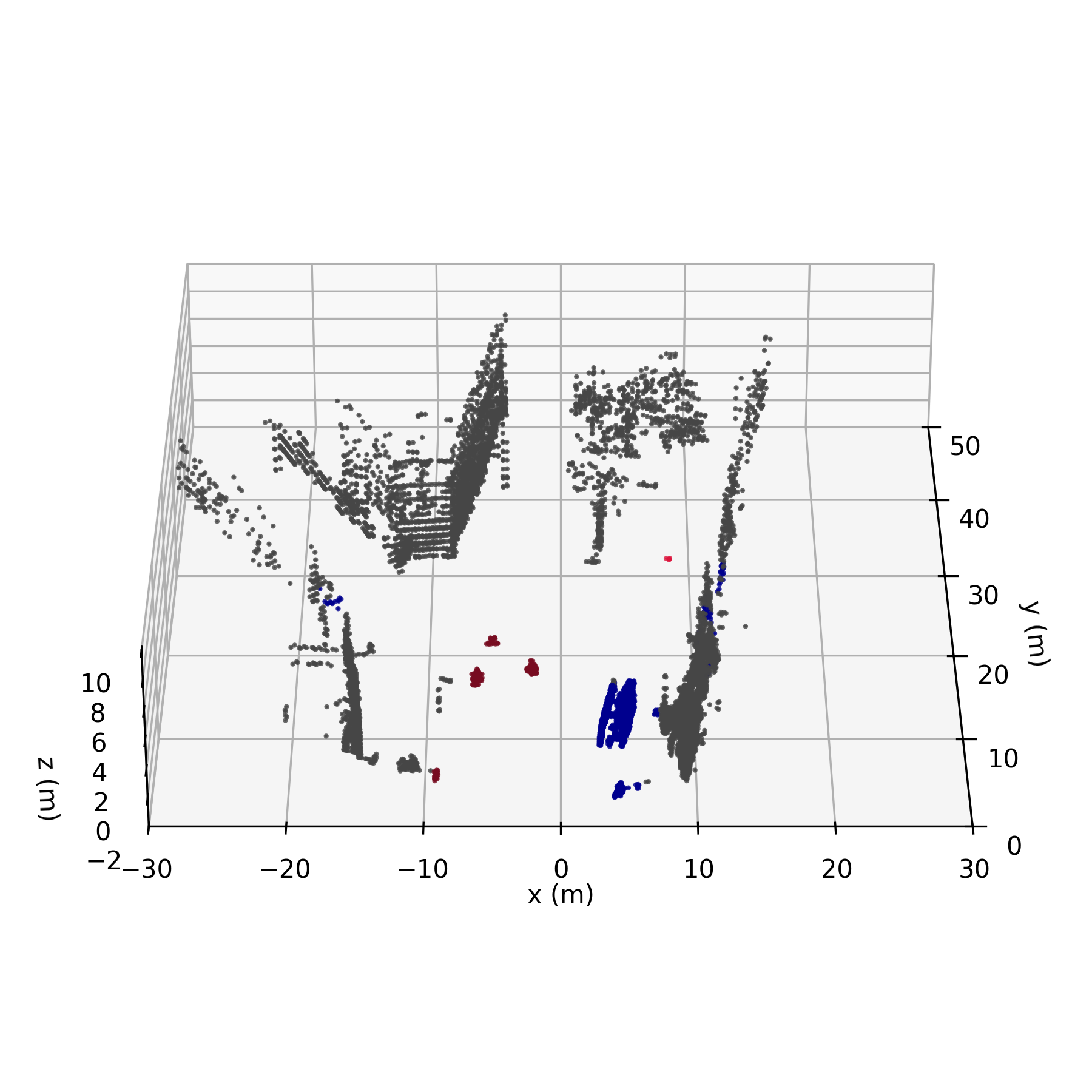}
  
  \end{subfigure}\hfill
  \begin{subfigure}{0.48\linewidth}
    \caption{Generated radar PCs}
    \includegraphics[width=\linewidth, trim = 0mm 30mm 0mm 30mm, clip]{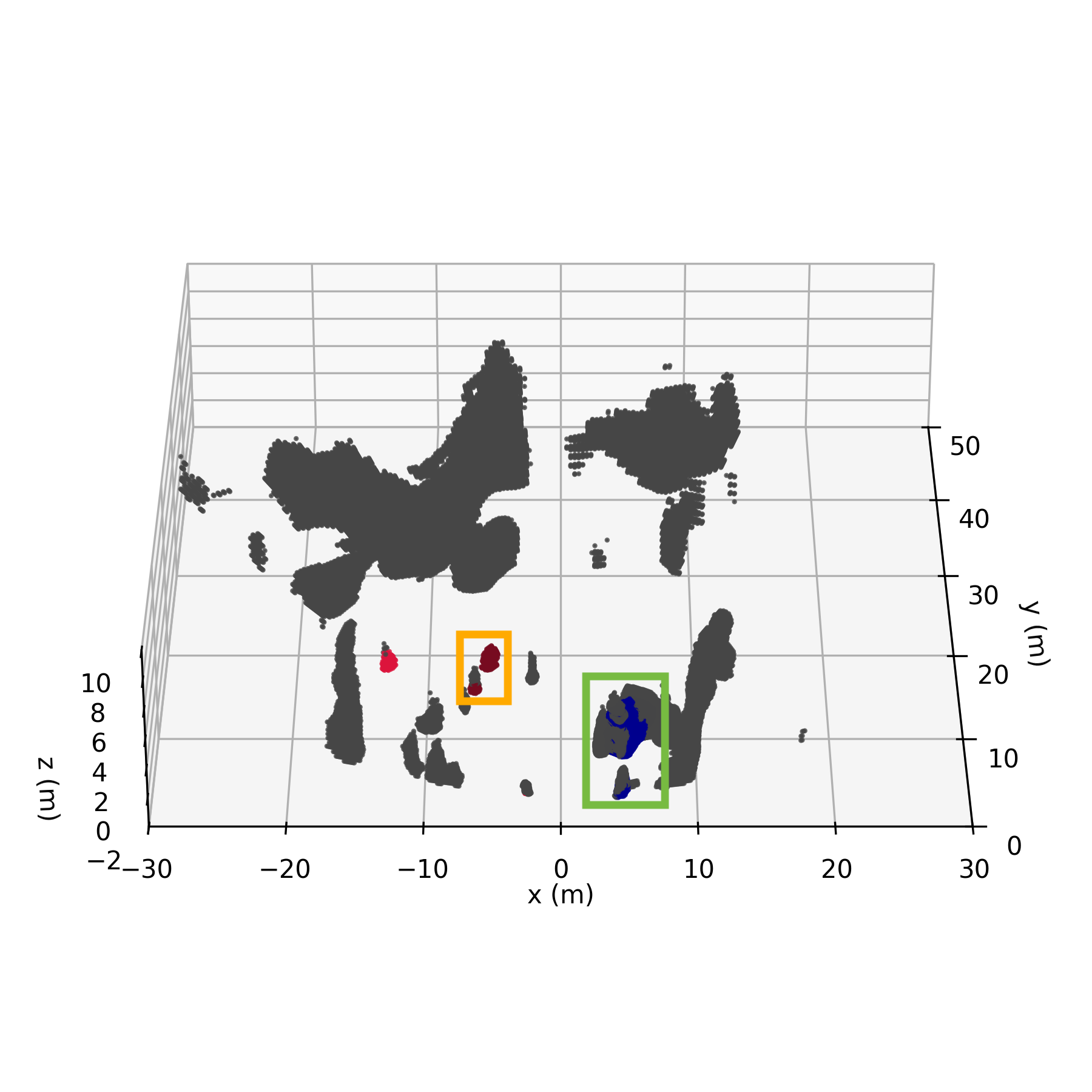}
   
  \end{subfigure}

  \caption{Visualization of generated radar PCs and the corresponding image and LiDAR ground truth, different colors represents different classes. The color bar distinguishes the 4 classes. Overall, the generated radar PCs correctly identify the vehicles, as in green box, and bicycles, as in orange box.}
  \label{fig:reproduce}
\end{figure}

\subsection{Camera-Assisted Labeling Performance}
To assess the effectiveness of camera-based and camera-radar-fused labeling under varying visual degradation, we evaluated the detection probability ($P_d$), false alarm probability ($P_{fa}$) and chamfer distance (CD) in increasing fog intensities. Table \ref{tab:fog_results} summarizes the performance for the camera-only, camera + radar and radar-only configurations, compared to the baseline RaDelft method. All methods are evaluated in Scene2 of the RaDelft dataset using the first 150 frames, where radar predictions are projected into the camera frame and cropped in a depth range of $50m$ as in Equation \ref{eq:mask_valid}. The "RaDelft baseline" refers to our reproduced version of the original RaDelft model \cite{sun2025automatic}, evaluated under the same spatial and temporal conditions to ensure a fair comparison with other methods. The camera-only method uses semantic segmentation from the camera images alone to assign labels to radar points. The camera + radar method uses the fused camera and radar segmentation output to assign labels to radar points. The 'radar only' method relies solely on radar semantic segmentation results for labeling.

\begin{table*}[htbp]
\centering
\caption{Camera-assisted labeling performance under different fog intensities.}
\label{tab:fog_results}
\renewcommand{\arraystretch}{1.1}
\setlength{\tabcolsep}{4pt} 
\begin{tabular}{c l c c c c c c c c}
\toprule
\textbf{Fog Intensity} &
\textbf{Method} &
$P_{d\_\mathrm{Scenario}}$ &
$P_{fa\_\mathrm{Scenario}}$ &
$P_{d\_\mathrm{Vehicles}}$ &
$P_{fa\_\mathrm{Vehicles}}$ &
$P_{d\_\mathrm{VRU}}$ &
$P_{fa\_\mathrm{VRU}}$ &
$CD_\mathrm{Scenario}$ &
$CD_\mathrm{Target}$ \\
\midrule
0.00 & RaDelft baseline & 0.556 & 0.0155 & 0.331 & 0.00073 & 0.325 & 0.00025 & 1.87 & 6.30 \\
     & Camera only    & 0.541 & 0.0145 & 0.567 & 0.00179 & 0.426 & 0.00012 & 1.79 & 4.66 \\
\midrule
0.02 & Camera only    & 0.540 & 0.0145 & 0.566 & 0.00177 & 0.433 & 0.00012 & 1.79 & 4.79 \\
     & Camera+radar   & 0.535 & 0.0145 & 0.555 & 0.00176 & 0.426 & 0.00021 & 1.81 & 4.77 \\
\midrule
0.04 & Camera only    & 0.541 & 0.0145 & 0.564 & 0.00177 & 0.419 & 0.00012 & 1.79 & 4.93 \\
     & Camera+radar   & 0.535 & 0.0145 & 0.558 & 0.00178 & 0.418 & 0.00021 & 1.80 & 4.74 \\
\midrule
0.08 & Camera only    & 0.544 & 0.0147 & 0.544 & 0.00166 & 0.365 & 0.00010 & 1.79 & 5.69 \\
     & Camera+radar   & 0.537 & 0.0145 & 0.554 & 0.00173 & 0.377 & 0.00019 & 1.80 & 4.91 \\
\midrule
0.15 & Camera only    & 0.548 & 0.0151 & 0.463 & 0.00136 & 0.059 & 0.00002 & 1.81 & 6.94 \\
     & Camera+radar   & 0.540 & 0.0147 & 0.523 & 0.00157 & 0.329 & 0.00017 & 1.79 & 5.23 \\
\midrule
--   & Radar only     & 0.544 & 0.0153 & 0.357 & 0.00095 & 0.330 & 0.00023 & 1.87 & 5.54 \\
\bottomrule
\end{tabular}
\end{table*}

In clear weather, the reproduced RaDelft baseline achieves a moderate detection rate ($P_{d\_\mathrm{Vehicles}} = 0.33$) with very low false alarms, but it tends to miss many small objects. By contrast, the camera-only method significantly improves the detection of vehicles and VRUs while maintaining a comparable false alarm rate and achieving a lower Chamfer Distance, indicating a more accurate labeling of radar point clouds. As the fog intensity increases, the performance of the camera-only labeling gradually degrades, particularly for vulnerable road users (VRUs). The fused camera+radar method consistently improves the geometric accuracy (lower $CD_{\mathrm{Target}}$) and partially improves the object detection rates compared to the camera-only baseline, especially at heavy fog levels (e.g., fog intensity $0.15$). This demonstrates that radar contributes complementary information that remains robust under adverse visual conditions. The radar-only model, trained purely on radar depth maps, shows stable but limited performance. Overall, the results show that camera-assisted labeling provides better performance than the original RaDelft baseline methods.

\begin{figure}[htbp]
  \centering

\begin{subfigure}{0.48\linewidth}
    \caption{(a) Corresponding image}
    \includegraphics[width=\linewidth, trim = 0mm 30mm 0mm 30mm, clip]{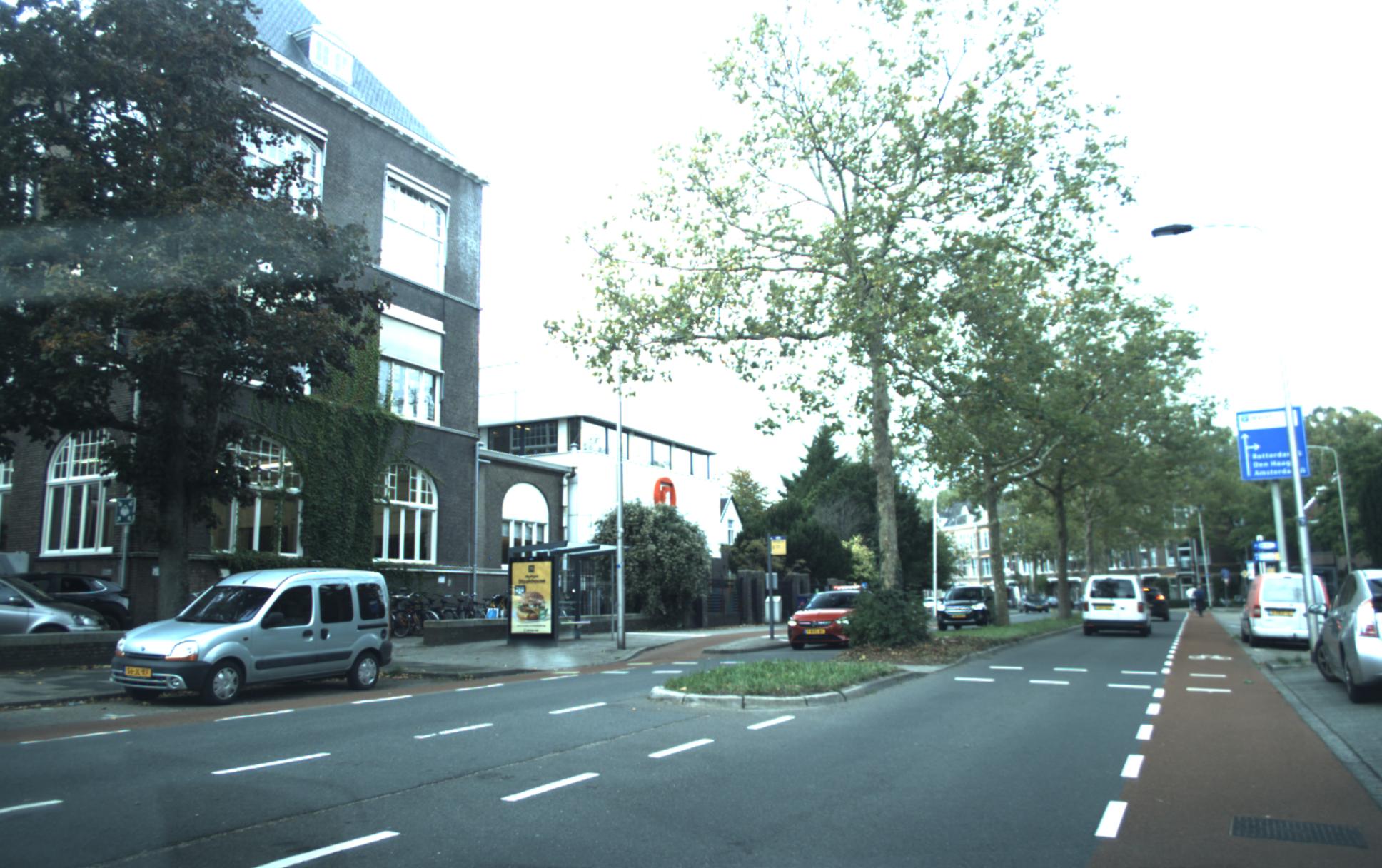}
  
  \end{subfigure}\hfill
  \begin{subfigure}{0.48\linewidth}
    \caption{(b) LiDAR ground truth}
    \includegraphics[width=\linewidth, trim = 0mm 30mm 0mm 30mm, clip]{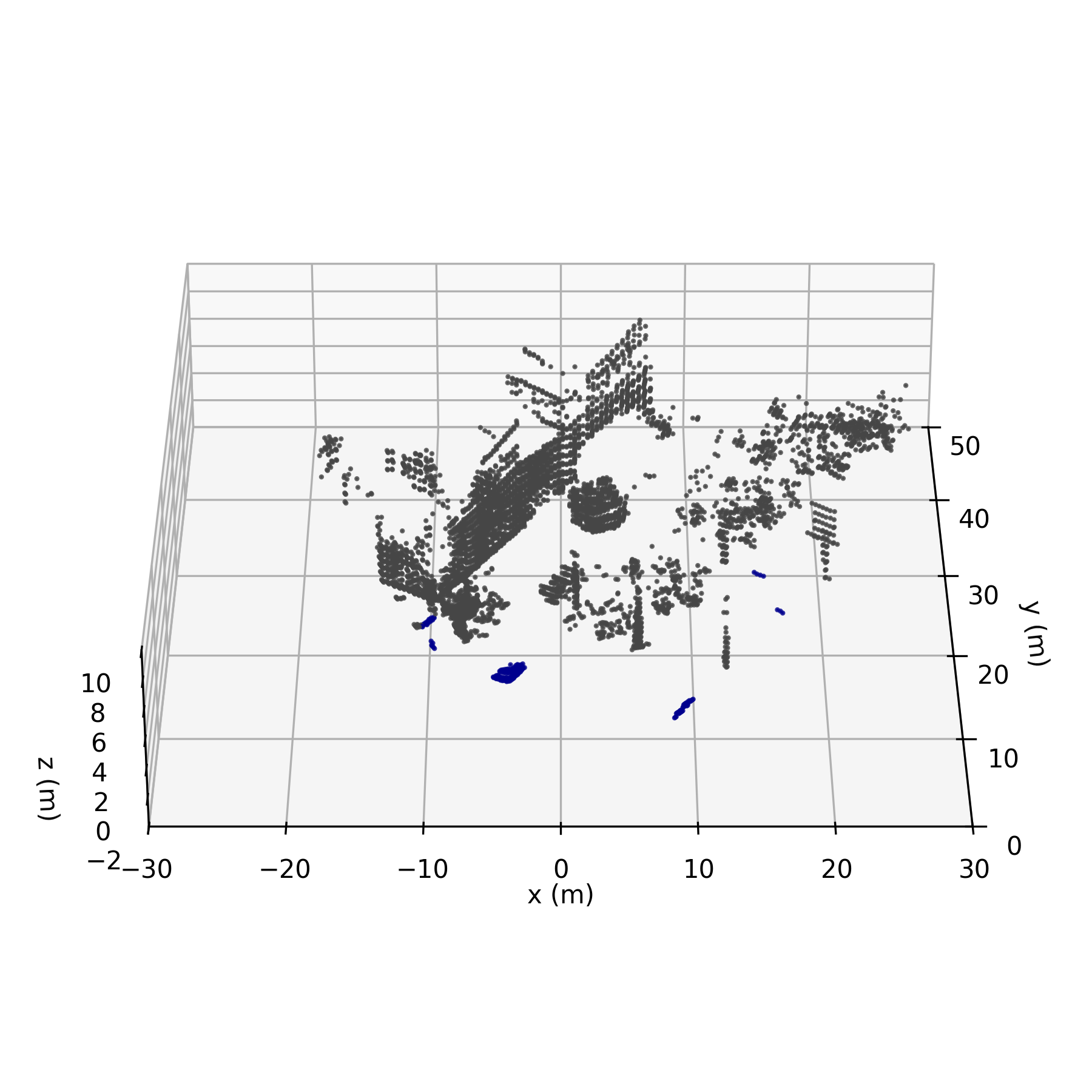}
   
  \end{subfigure}

  \vspace{0.4em}

  \begin{subfigure}{0.48\linewidth}
    \caption{(c) Reproduced RaDelft radar PCs}
    \includegraphics[width=\linewidth, trim = 0mm 30mm 0mm 30mm, clip]{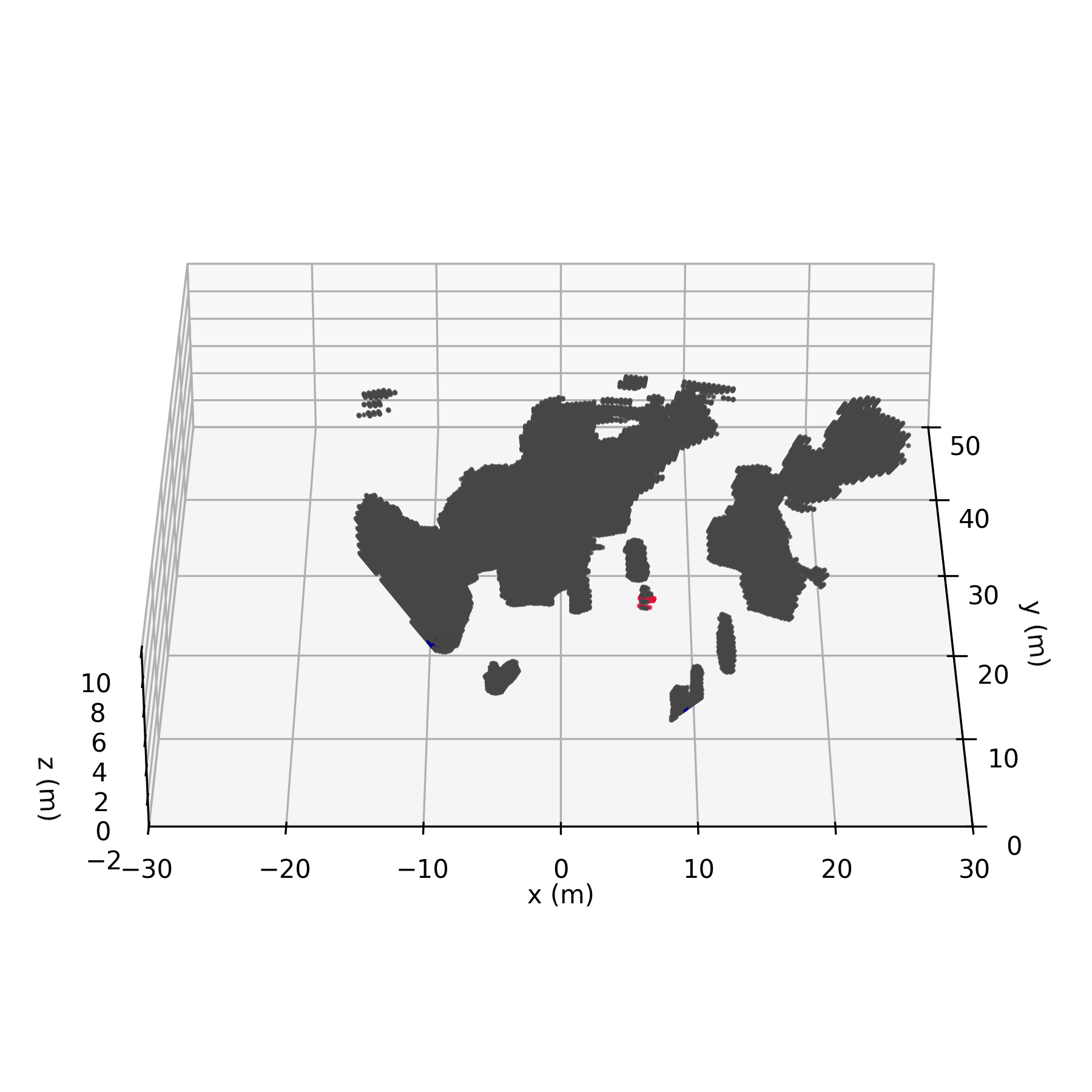}
  
  \end{subfigure}\hfill
  \begin{subfigure}{0.48\linewidth}
    \caption{(d) Camera-Assisted relabeled radar PCs}
    \includegraphics[width=\linewidth, trim = 0mm 30mm 0mm 30mm, clip]{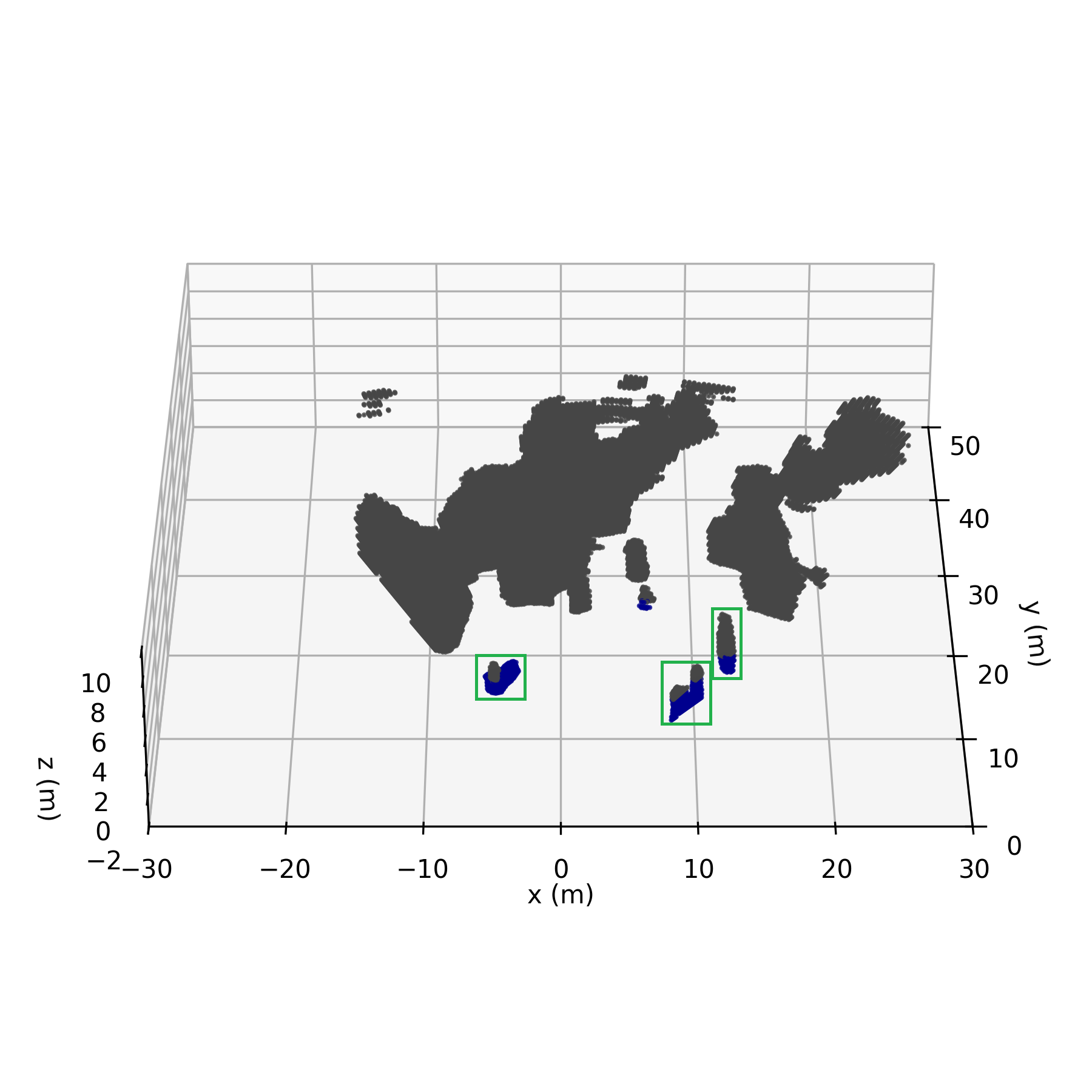}
   
  \end{subfigure}

  \caption{Qualitative comparison of radar point cloud labeling. Our method, in figure (d), correctly classified the cars (points within the green box). Whereas the original RaDelft method failed to classify the vehicles.}
  \label{fig:label_comparison}
\end{figure}

Fig.~\ref{fig:label_comparison} provides a qualitative comparison between the reproduced RaDelft baseline labeling and our camera-assisted labeling approach (camera-only variant). The top row shows the RGB camera image and the corresponding LiDAR point cloud used as a geometric reference. The bottom row shows the radar point cloud colored by semantic labels produced by the reproduced RaDelft model in (c); and our camera-derived model in (d). Our method produces denser and more consistent labels on vehicles compared to the reproduced RaDelft output.

\section{Discussion}
Although the fusion of radar and camera segmentation generally improves the performance in extreme adverse conditions (e.g., fog intensity $\beta = 0.15$), our results indicate that under light fog (e.g., fog intensity $\beta = 0.02$ and $\beta = 0.04$), the "camera + radar" approach does not outperform the "camera only" labeling. In these conditions, the camera segmentation still provides high-confidence semantic results, and the added radar information has limited complementary value. Moreover, the radar depth maps have lower spatial resolution. When fused, these coarse radar predictions can introduce additional false positives in regions where the camera segmentation is already accurate, slightly degrading $P_{fa}$. This suggests that multi-modal fusion should be adaptively weighted based on environmental visibility, where camera-dominant labeling suffices in light fog, while radar contributions become more important only as visual quality deteriorates.
\section{Conclusions}
In this work, we reproduced the RaDelft 4D radar semantic segmentation baseline, providing an open, reproducible framework including the full training pipeline, pretrained weights, and labeled radar data. We further introduced a camera-assisted labeling approach that leverages camera semantic segmentation to generate accurate radar labels through geometric projection and spatial clustering. Our experiments demonstrate that this approach significantly improves labeling accuracy compared to the original RaDelft baseline and provides a useful benchmark for future radar perception research. We also analyzed the robustness of camera-based supervision under varying fog intensities. The results show that camera-only labeling performs well under clear and light fog conditions, and fusion with radar information only becomes useful when the visibility deteriorates badly. Further research can focus on developing adaptive fusion strategies that dynamically adjust the weighting between camera and radar modalities based on environmental conditions. Moreover, we can incorporate temporal information for motion-consistent labeling.

\appendices

\bibliographystyle{IEEEtran}
\bibliography{references}

\end{document}